\documentclass[journal]{IEEEtran}

\usepackage{times}
\usepackage{mathrsfs}
\usepackage{epsfig}
\usepackage{graphicx}
\usepackage{amsmath}
\usepackage{amssymb}
\usepackage{graphicx}
\usepackage{cite}
\usepackage{enumerate}
\usepackage{cases}
\usepackage{multirow}
\usepackage{verbatim}
\usepackage{amssymb}
\usepackage{CJK}
\usepackage{algorithm}
\usepackage{algorithmicx}
\usepackage{algpseudocode}
\usepackage{color}
\usepackage{bm}
\usepackage{booktabs}
\usepackage{amssymb}

\usepackage[pagebackref=true,breaklinks=true,letterpaper=true,colorlinks,bookmarks=false]{hyperref}

\usepackage{color}
\usepackage{tikz}
\usepackage{ifthen}
\usepackage{bm}
\usetikzlibrary{3d,decorations.text,shapes.arrows,positioning,fit,backgrounds, intersections}
\usetikzlibrary{shapes.misc}
\definecolor{rdcolor}{RGB}{171, 197, 196}
\definecolor{ddcolor}{RGB}{173, 185, 202}

\definecolor{color3}{RGB}{249, 205, 173}
\definecolor{color4}{RGB}{229, 131, 8}
\definecolor{color7}{HTML}{BF7130}
\definecolor{color9}{HTML}{A5B92E}

\definecolor{goldenrod}{RGB}{85, 65, 13}
\definecolor{navyblue}{RGB}{0, 0, 50}
\definecolor{dandelion}{RGB}{94, 88, 19}
\definecolor{brickred}{RGB}{80, 25, 33}
\definecolor{rfcolor}{HTML}{FF4500}
\definecolor{dfcolor}{HTML}{61D8A2}
\definecolor{gmacolor}{RGB}{237, 147, 147}
\definecolor{gmaedge}{RGB}{227, 91, 91}
\definecolor{fcolor}{RGB}{251, 241, 117}
\definecolor{fedge}{RGB}{217, 212, 0}
\definecolor{mfcolor}{RGB}{255, 116, 0}
\definecolor{mfEdge}{RGB}{237, 107, 55}
\definecolor{linecolor}{RGB}{16, 73, 94}
\tikzset{drawLine/.style={->,  line width=1.5pt, color=linecolor!55}}
\tikzset{rfLine/.style={->, line width=1.5pt,color=rfcolor!50}}
\tikzset{dfLine/.style={->, line width=1.5pt,color=dfcolor!50}}
\definecolor{edgeColor}{RGB}{32, 32, 32}
\tikzset{circle dotted/.style={dash pattern=on .05mm off 2mm,
		line cap=round}}
\tikzset{global scale/.style={
		scale=#1,
		every node/.append style={scale=#1}
	}}

\tikzstyle{GMASTYLE}=[draw, rectangle, color=gmaedge, text=black, minimum height=6mm, minimum width=7mm/0.618, fill=gmacolor!60, rounded corners=4pt, line width=.5pt]
\tikzstyle{FSTYLE}=[draw, circle,color=fedge, minimum width=4mm, inner sep=0.1mm, fill=fcolor!60,  text=black, scale=0.9]
\tikzstyle{SSTYLE}=[draw, circle, minimum size=12pt, inner sep=0pt, fill=navyblue!30]
\tikzstyle{ASTYLE}=[draw, circle, minimum size=12pt, inner sep=0pt, fill=pink!30]
\tikzstyle{MFSTYLE}=[draw, rounded rectangle, color=mfEdge, minimum height=6mm,fill=mfcolor!45, text=black]
\usepackage{tkz-euclide}
\usetkzobj{all}
\tikzset{
	connect/.style args={(#1) to (#2) over (#3) by #4}{
		insert path={
			let \p1=($(#1)-(#3)$), \n1={veclen(\x1,\y1)},
			\n2={atan2(\y1,\x1)}, \n3={abs(#4)}, \n4={#4>0 ?180:-180}  in
			(#1) -- ($(#1)!\n1-\n3!(#3)$)
			arc (\n2:\n2+\n4:\n3) -- (#2)
		}
	},
}
\tikzset{dist/.style={path picture= {
			\begin{scope}[x=1pt,y=10pt]
				\draw plot[domain=-6:6] (\x,{1/(1 + exp(-\x))-0.5});
			\end{scope}
		}
	}
}
				\newcommand{\networkLayer}[8]{
					\def\a{#1} 
					\def\b{0.02}
					\def\c{#2} 
					\def\t{#3} 
					\def\d{#4} 
					\def\z{#5} 
					\def\round{0.4pt}
					\def\linept{2pt}
					
					\ifthenelse{\equal{#8}{}}{
						\draw[line width=\linept, rounded corners=\round, color=edgeColor](\c+\t,\z,\d) -- (\c+\t,\a+\z,\d) --(\t,\a+\z,\d); 	
					}{
					\draw[line width=\linept, rounded corners=\round, color=edgeColor](\c+\t,\z,\d) -- (\c+\t,\a+\z,\d) --(\t,\a+\z,\d)node[midway, above]{#8}; 				
				} 
				\draw[line width=\linept, rounded corners=\round, color=edgeColor](\t,\z,\a+\d) -- (\c+\t,\z,\a+\d) node[midway,below] {#7} -- (\c+\t,\a+\z,\a+\d) -- (\t,\a+\z,\a+\d)  -- (\t,\z,\a+\d) --cycle; 
				\draw[line width=\linept, rounded corners=\round, color=edgeColor](\c+\t,\z,\d) -- (\c+\t,\z,\a+\d) --(\c+\t, \z+\a, \a+\d) -- (\c+\t, \a+\z, \d)--cycle;
				\draw[line width=\linept, rounded corners=\round, color=edgeColor](\c+\t,\a+\z,\d) -- (\c+\t,\a+\z,\a+\d);
				\draw[line width=\linept, rounded corners=\round, color=edgeColor](\t, \a+\z, \d) -- (\t, \a+\z, \a+\d)-- (\c+\t, \a+\z, \a+\d)-- (\t+\c, \a+\z, \d)--cycle; 

				\filldraw[#6,line width=0pt,rounded corners=\round] (\t+\b,\b+\z,\a+\d) -- (\c+\t-\b,\b+\z,\a+\d) -- (\c+\t-\b,\a-\b+\z,\a+\d) -- (\t+\b,\a-\b+\z,\a+\d) ; 
				
				\filldraw[#6,line width=0pt,rounded corners=\round] (\t+\b,\a+\z,\a-\b*4+\d) -- (\c+\t-\b*2,\a+\z, \a-\b*4+\d) -- (\c+\t-\b*2,\a+\z,\b+\d) -- (\t+\b,\a+\z,\b+\d); 
				\ifthenelse {\equal{#6} {}}
				{} 
				{\filldraw[#6,line width=0pt, rounded corners=\round] (\c+\t,\b*2+\z,\a-\b*4+\d-\b) -- (\c+\t,\b*2+\z,\b+\d) -- (\c+\t,\a-\b*2.68+\z,\b+\d) -- (\c+\t,\a-\b*2.68+\z,\a-\b*4+\d-\b);} 
			}
\usepackage{multirow}
\def\metrics{$F_{\beta}\uparrow$\quad $S_m \uparrow$\quad MAE $\downarrow$ }
\def\triplets(#1,#2,#3){#1\quad#2\quad#3}
\def\fillcell{$\text{--}\phantom{0.8888}\text{--}\phantom{0.8888}\text{--}$}
\usepackage{enumitem}
\def\OURNET{\textit{DPANet}}
\usepackage{booktabs}

\newcommand{\etal}{\textit{et al}.}
\newcommand{\ie}{\textit{i}.\textit{e}.}
\newcommand{\eg}{\textit{e}.\textit{g}.}
\newcommand{\etc}{\textit{etc}.}

\begin{document}

\title{DPANet: Depth Potentiality-Aware Gated Attention Network for RGB-D Salient Object Detection}

\author{Zuyao Chen, Runmin Cong,~\IEEEmembership{Member,~IEEE,} Qianqian Xu,~\IEEEmembership{Senior Member,~IEEE,}\\ and Qingming Huang,~\IEEEmembership{Fellow,~IEEE}
\thanks{Manuscript received Mar. 2020. This work was supported in part by the National Key R\&D Program of China under Grant 2018AAA0102003, in part by the Beijing Nova Program under Grant Z201100006820016, in part by the National Natural Science Foundation of China under Grant 62002014, Grant 61620106009, Grant U1636214, Grant 61931008, Grant 61836002, Grant 61672514, Grant 61976202, in part by Key Research Program of Frontier Sciences under Grant CAS: QYZDJ-SSW-SYS013, in part by the Strategic Priority Research Program of Chinese Academy of Sciences under Grant XDB28000000, in part by Beijing Natural Science Foundation under Grant 4182079, and in part by Youth Innovation Promotion Association CAS, in part by Hong Kong Scholars Program, in part by China Postdoctoral Science Foundation under Grant 2020T130050, Grant 2019M660438, and in part by the Fundamental Research Funds for the Central Universities under Grant 2019RC039.  (\emph{Zuyao Chen and Runmin Cong contributed equally to this work.}) (\emph{Corresponding authors: Qianqian Xu and Qingming Huang.})
}
\thanks{Z. Chen is with the School of Computer Science and Technology, University of Chinese Academy of Sciences, Beijing 100190, China (e-mail: chenzuyao17@mails.ucas.ac.cn).}
\thanks{R. Cong is with the Institute of Information Science, Beijing Jiaotong University, Beijing 100044, China, also with the Beijing Key Laboratory of Advanced Information Science and Network Technology, Beijing 100044, China, and also with the Department of Computer Science, City University of Hong Kong, Hong Kong SAR, China (e-mail: rmcong@bjtu.edu.cn).}
\thanks{Q. Xu is with the Key Laboratory of Intelligent Information Processing, Institute of Computing Technology, Chinese Academy of Sciences, Beijing 100190, China (e-mail: xuqianqian@ict.ac.cn).}
\thanks{Q. Huang is with the School of Computer Science and Technology, University of Chinese Academy of Sciences, Beijing 101408, China, also with the Key Laboratory of Big Data Mining and Knowledge Management (BDKM), University of Chinese Academy of Sciences, Beijing 101408, China, also with the Key Laboratory of Intelligent Information Processing, Institute of Computing Technology, Chinese Academy of Sciences, Beijing 100190, China, and also with Peng Cheng Laboratory, Shenzhen 518055, China (e-mail: qmhuang@ucas.ac.cn).}
}

\markboth{IEEE TRANSACTIONS ON IMAGE PROCESSING}
{Shell \MakeLowercase{\textit{et al.}}: Bare Demo of IEEEtran.cls for IEEE Journals}
\maketitle

\begin{abstract}
There are two main issues in RGB-D salient object detection: (1) how to effectively integrate the complementarity from the cross-modal RGB-D data; (2) how to prevent the contamination effect from the unreliable depth map. In fact, these two problems are linked and intertwined, but the previous methods tend to focus only on the first problem and ignore the consideration of depth map quality, which may yield the model fall into the sub-optimal state. In this paper, we address these two issues in a holistic model synergistically, and propose a novel network named \OURNET\ to explicitly model the potentiality of the depth map and effectively integrate the cross-modal complementarity. By introducing the depth potentiality perception, the network can perceive the potentiality of depth information in a learning-based manner, and guide the fusion process of two modal data to prevent the contamination occurred.\ The gated multi-modality attention module in the fusion process exploits the attention mechanism with a gate controller to capture long-range dependencies from a cross-modal perspective.\ Experimental results compared with 16 state-of-the-art methods on 8 datasets demonstrate the validity of the proposed approach both quantitatively and qualitatively. \url{https://github.com/JosephChenHub/DPANet}
\end{abstract}

\begin{IEEEkeywords}
Salient object detection, RGB-D images, Depth potentiality perception, Gated multi-modality attention.
\end{IEEEkeywords}

\IEEEpeerreviewmaketitle

\section{Introduction} \label{sec1}
\begin{figure}[t]
 	\centering
 	\includegraphics[width=\columnwidth]{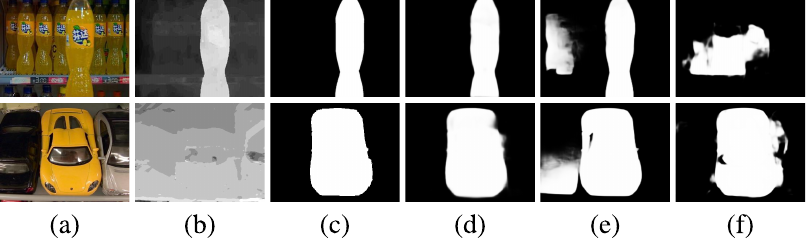}
 	\caption {Sample results of our method compared with others. RGB-D methods are marked in \textbf{boldface}.
 		(a) RGB image; (b) Depth map; (c) Ground truth; (d) \textbf{Ours}; (e) BASNet \cite{Qin_2019_CVPR}; (f) \textbf{CPFP} \cite{zhao2019contrast}.
 	}
 	\label{fig1}
 \end{figure}

\IEEEPARstart{S}{ALIENT} object detection (SOD) aims to locate interesting regions that attract human attention most in an image \cite{Survey, REVIEW}. As a pre-processing technique, SOD benefits a variety of applications including object segmentation \cite{R1}, person re-identification \cite{zhao2013unsupervised}, image understanding \cite{zhang2014saliency}, thumbnail creation \cite{R2}, image quality assessment \cite{R3}, and image enhancement \cite{R4}. In the past years, CNN-based methods have achieved promising performances in the SOD task owing to its powerful representation ability of CNN \cite{krizhevsky2012imagenet}. Most of them \cite{DSS,deng2018r3net,liu2018picanet,Liu2019PoolSal,Qin_2019_CVPR,zhao2019EGNet,feng2019attentive,PFANet,GCPANet,Unsupervised,weaksupervised,wang2013saliency,wang2018detecting,nips20} focused on detecting the salient objects from the RGB image, while it is hard to achieve better performance in some challenging and complex scenarios using only one single modal data, such as similar appearance between the foreground and background (see the first row in Fig.\ \ref{fig1}), the cluttered background interferences (see the second row in Fig.\ \ref{fig1}).\ Recently, depth information has become increasingly popular thanks to the affordable and portable devices, \eg, Microsoft Kinect and iPhone XR, which can provide many useful and complementary cues in addition to the color appearance information, such as shape structure and boundary information and has been successfully applied in many vision tasks \cite{crmspl,choi13iros_rgbdtracking,crmtip18,crmtc19,ijcai20,crmtmm19,crmtc20,lcy2020tc,eccv20}. Introducing depth information into SOD does address these challenging scenarios to some degree. However, as shown in Fig. \ref{fig1}, there exists a conflict that depth maps are sometimes inaccurate and would contaminate the results of SOD. Previous works generally integrate the RGB and depth information in an indiscriminate manner, which may induce negative results when encountering the inaccurate or blurred depth maps. Moreover, it is often insufficient  to fuse and capture complementary information of different modal from the RGB image and depth map via simple strategies such as cascading, multiplication, which may degrade the saliency result.
Hence, there are two main issues in RGB-D SOD to be addressed, \ie, 1) how to prevent the contamination from unreliable depth information; 2) how to effectively integrate the multi-modal information from the RGB image and corresponding depth map.
\\
\indent As a remedy for the above-mentioned issues, we propose a Depth Potentiality-Aware Gated Attention Network (\OURNET) that can simultaneously model the potentiality of the depth map, and optimize the fusion process of RGB and depth information in a gated attention mechanism. Instead of indiscriminately integrating multi-modal information from the RGB image and depth map, we focus on adaptively fusing two modal data by considering the depth potentiality perception in a learning-based manner. The depth potentiality perception works as a controller to guide the aggregation of cross-modal information and prevent the contamination from the unreliable depth map. For the depth potentiality perception, we take the saliency detection as the task orientation to measure the relationship between the binary depth map and the corresponding saliency mask. If the binary depth map obtained by a simple thresholding method (\eg, Ostu \cite{otsu1979threshold}) is close to the ground truth, the reliability of the depth map is high, and therefore a higher depth confidence response should be assigned for this depth input.
The depth confidence response will be adopted in the gated multi-modality fusion to better integrate the cross-modal information and prevent the contamination. \\
\indent Considering the complementarity and inconsistency of RGB and depth information, we propose a gated multi-modality attention (GMA) module to capture long-range dependencies from a cross-modal perspective. Concatenating or summing the cross-modal features from RGB-D images not only has information redundancy, but also makes the truly effective complementary information submerged in  a large number of data features. Therefore, the GMA module we designed utilizes the spatial attention mechanism to extract the most discriminative features, and can adaptively use the gate function to control the fusion rate of the cross-modal information and reduce the negative impact caused by the unreliable depth map. Moreover, we design a multi-level feature fusion mechanism to better integrate different levels of features including single-modality and multi-modality features. As is shown in Fig. \ref{fig1}, the proposed network can handle some challenging scenarios, such as the background disturbances (similar appearance), complex scenes, and unreliable depth maps. \\
\indent In summary, our main contributions are listed as follows:
\begin{itemize}[noitemsep, topsep=0pt]
	\item For the first time, we address the unreliable depth map in the RGB-D SOD network in an end-to-end formulation, and propose the \OURNET\ by incorporating the depth potentiality perception into the cross-modality integration pipeline.
	\item Without increasing the training label (\ie, depth quality label), we model a task-orientated depth potentiality perception module that can adaptively perceive the potentiality of the input depth map, and further weaken the contamination from unreliable depth information.
	\item We propose a GMA module to effectively aggregate the cross-modal complementarity of the RGB and depth images, where the spatial attention mechanism aims at reducing the information redundancy, and the gate function controller focuses on regulating the fusion rate of the cross-modal information.
	\item Without any pre-processing (\eg, HHA \cite{gupta2014learning}) or post-processing (\eg, CRF \cite{krahenbuhl2011efficient}) techniques, the proposed network outperforms $16$ state-of-the-art methods on $8$ RGB-D SOD datasets in quantitative and qualitative evaluations.
\end{itemize}
\section{Related Work}
In this section, we will review the salient object detection models for RGB and RGB-D images, especially the deep learning based methods, which have achieved impressive progress in recent years. \\

\subsection{RGB Salient Object Detection}
The last decades have witnessed the prosperous development and improvement of salient object detection for RGB image. Specifically, from the bottom-up models \cite{DSR,RBD,DCLC,SMD} to top-down models \cite{DSS,deng2018r3net,liu2018picanet,Liu2019PoolSal,Qin_2019_CVPR,zhao2019EGNet,feng2019attentive,PFANet,GCPANet,Unsupervised,weaksupervised}, the performance is constantly being refreshed and surpassed. The bottom-up models used some priors (\eg, background prior \cite{RBD}, compactness prior \cite{DCLC} ) and properties (\eg, sparse \cite{DSR}, low rank \cite{SMD}) to define the saliency. Recently, deep learning based SOD methods have gradually become mainstream and achieved significant performance improvements.
Hou \etal \cite{DSS} introduced short connections into the skip-layer structures within the holistically-nested edge detector network architecture to achieve accurate saliency detection.
The edge or boundary cue is embedded in the deep model to highlight the boundary region of salient object and improve the performance of SOD, such as EGNet \cite{zhao2019EGNet}, BASNet \cite{Qin_2019_CVPR}, and AFNet \cite{feng2019attentive}.
Attention mechanism has been successfully applied in SOD to learn more discriminative feature representation, such as PiCANet \cite{liu2018picanet}, PFANet \cite{PFANet}, and GCPANet \cite{GCPANet}.
In addition, some studies focus on real-time SOD or weakly-/un- supervised learning. Liu \etal \cite{Liu2019PoolSal} explored the role of pooling in neural networks for SOD, which can yield detail enriched saliency maps. Zhang \etal \cite{Unsupervised} used the noisy label to train the SOD model and achieved comparable performance with the state-of-the-art supervised deep models. Zeng \etal \cite{weaksupervised} proposed a unified framework to train SOD models with diverse weak supervision sources including category labels, captions, and unlabelled data.

\subsection{RGB-D Salient Object Detection}
For the unsupervised SOD method for RGB-D images, some handcrafted features are designed, such as depth contrast, depth measure. Peng \etal \cite{peng2014rgbd} proposed to fuse the RGB and depth at first and feed it to a multi-stage saliency model. Song \etal \cite{song2017depth}  proposed a multi-scale discriminative saliency fusion framework. Feng \etal \cite{feng2016local} proposed a Local Background Enclosure (LBE) to capture the spread of angular directions.
Ju \etal \cite{ju2014depth} proposed a depth-induced method based on using anisotropic center-surround difference. Fan \etal \cite{fan2014salient} combined the region-level depth, color and spatial information to achieve saliency detection in stereoscopic images. Cheng \etal \cite{cheng2014depth} measured the salient value using color contrast, depth contrast, and spatial bias. \\
\indent Lately, deep learning based approaches have gradually become a mainstream trend in RGB-D saliency detection. Qu \etal \cite{qu2017rgbd} proposed to fuse different low-level saliency cues into hierarchical features, including local contrast, global contrast, background prior and spatial prior. Zhu \etal \cite{zhu2019pdnet} designed a master network to process RGB values, and a sub-network for depth cues and incorporate the depth-based features into the master network. Chen \etal \cite{chen2018progressively} designed a progressive network attempting to integrate cross-modal complementarity. Zhao \etal \cite{zhao2019contrast} integrated the RGB features and enhanced depth cues using depth prior for SOD. Piao \etal \cite{Piao_2019_ICCV} proposed a depth-induced multi-scale recurrent attention network. However, these efforts attempting to integrate the RGB and depth information indiscriminately ignore contaminations from inaccurate or blurred depth maps. Fan \etal \cite{fan2019D3Net} attempted to address this issue by designing a depth depurator unit to abandon the low-quality depth maps.
\begin{figure*}[t]
	 \centering
	\includegraphics[width=0.98\textwidth]{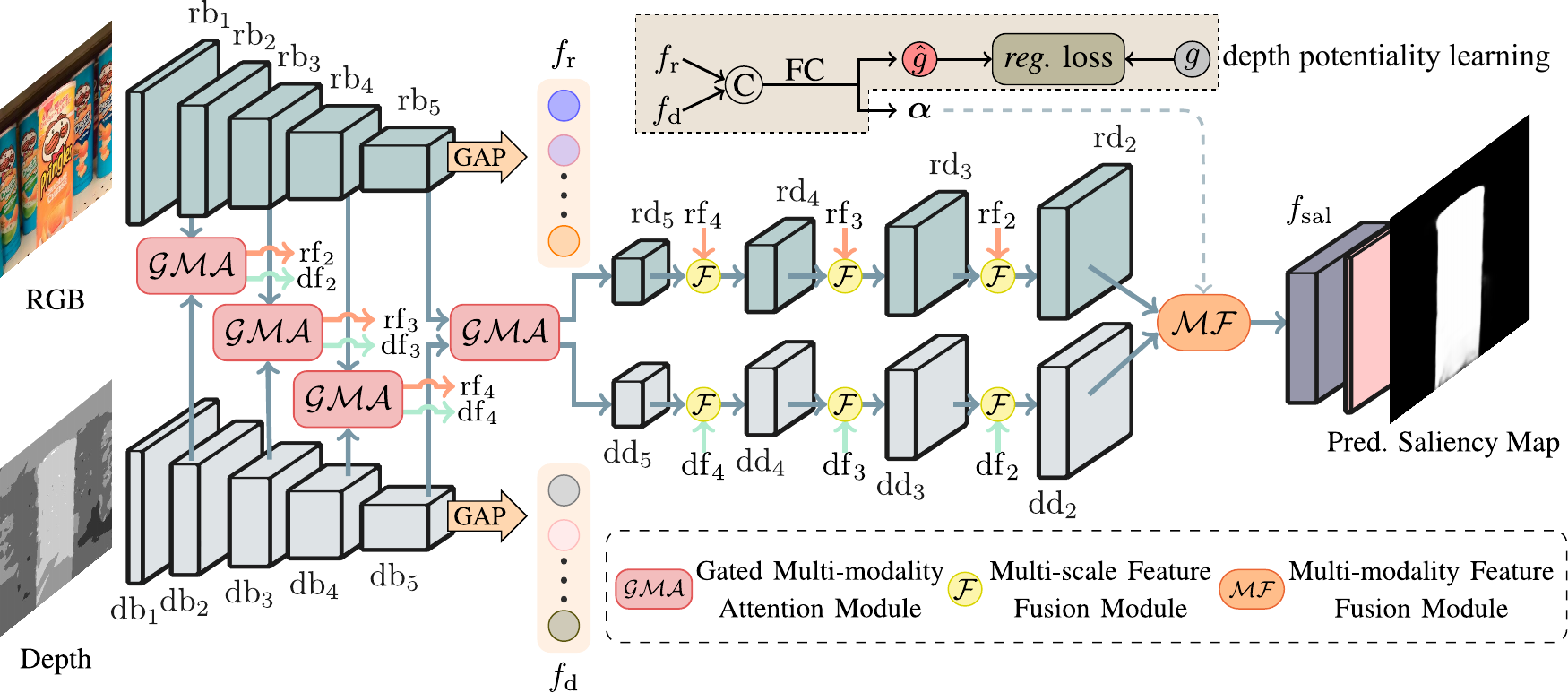}	
\caption{Architecture of \OURNET. For better visualization, we only display the modules and features of each stage. $\mathrm{rb}_i$, $\mathrm{db}_i$ ($i=1,2,\cdots,5$)	denote the features generated by the backbone of the two branches respectively, and $\mathrm{rd}_i$, $\mathrm{dd}_i$ ($i=5,4,3,2$) represent the features of decoder stage. $\mathrm{rf}_i, \mathrm{df}_i$ ($i=2,3,4,5$, $\mathrm{rf}_5=\mathrm{rd}_5$, $\mathrm{df}_5=\mathrm{dd}_5$) refer to the output of the GMA module. $f_{\mathrm{sal}}$ is the generated final saliency map. ``C'' and ``FC'' refer to the concatenation and fully-connected layers respectively, and ``GAP'' represents the global average pooling.}
	 \label{fig:main_plot}	
\end{figure*}
\section{Methodology}
\subsection{Overview of the Proposed Network}
As shown in Fig. \ref{fig:main_plot}, the proposed network is a symmetrical two-stream encoder-decoder architecture. To be concise, we denote the output features of RGB branch in the encoder component as $\mathrm{rb}_i$ ($i=1,2,3,4,5$), and the features of depth branch in the encoder component as $\mathrm{db}_i$ ($i=1,2,3,4,5$). The feature $\mathrm{rb}_i$ ($i=2,3,4,5$) and feature $\mathrm{db}_i$ ($i=2,3,4,5$) are fed into a GMA module to obtain the corresponding enhanced feature $\mathrm{rf}_i$, $\mathrm{df}_i$, respectively. In GAM module, the weight of the gate is learned by the network in a supervised way. Specifically, the top layers' feature $\mathrm{rb}_5$ and $\mathrm{db}_5$ are passed through a global average pooling (GAP) layer and two fully connected layers to learn the predicted score of depth potentiality via the regression loss with the help of the pseudo labels.
Then, the decoder of two branches integrates multi-scale features progressively. Finally, we aggregate the two decoders' output and generate the saliency map by using the Multi-scale and Multi-modality Feature Fusion Modules. To facilitate the optimization, we add auxiliary loss branches at each sub-stage, \ie, $\mathrm{rd}_i$ and $\mathrm{dd}_i$ ($i=5,4,3,2$).
\subsection{Depth Potentiality Perception}
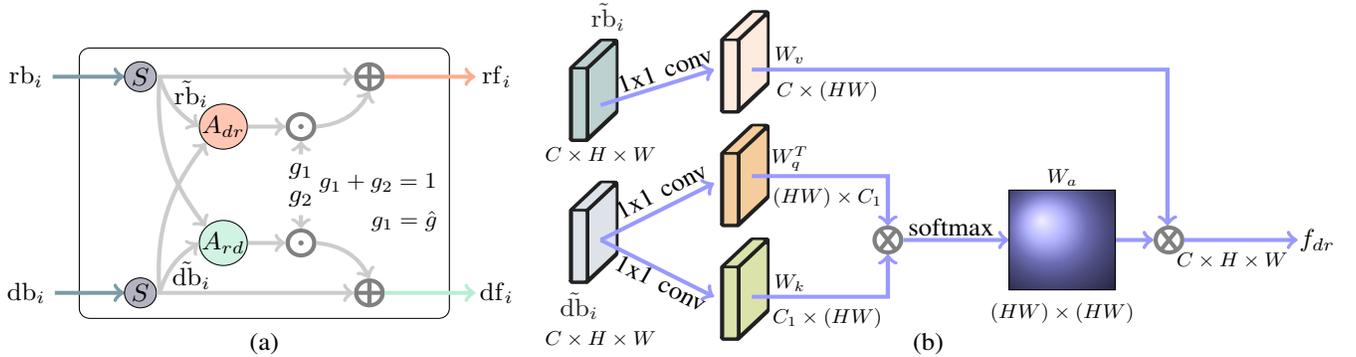
\begin{figure*}[t]
	\resizebox{\textwidth}{0.2\textheight}
	{
		\centering
		\begin{tikzpicture}[x=10mm, y=10mm]
		\node [draw, rectangle, rounded corners=4pt, minimum width=0.27\textwidth, minimum height=0.21\textwidth, ](b1)at (-0.45, 0) {};
		\node [] (rbi) at (-3.6, 1.5) {$\mathrm{rb}_i$};
		\node [below = 2.5 of rbi] (dbi) {$\mathrm{db}_i$};
		\draw [->, ultra thick, color=linecolor!55] (rbi) -- ++(1.3, 0) coordinate(p1);
		\draw [->, ultra thick, color=linecolor!55] (dbi) -- ++(1.3, 0) coordinate(p2);
		
		\node [SSTYLE] at ([xshift=6]p1) (s1)  {$S$} ;
		\node [SSTYLE] at ([xshift=6]p2) (s2) {$S$} ;
		
		\node [draw, ultra thick, circle, minimum width=10pt, color=gray] (c1) at ([xshift=80]s1.east) {};
		\draw [->, ultra thick, color=black!20] (s1.east) --(c1.west) coordinate(p3);
		
		\draw [ultra thick, color=gray] ([xshift=1]c1.west) -- ([xshift=-1]c1.east);
		\draw [ultra thick, color=gray] ([yshift=1]c1.south) -- ([yshift=-1]c1.north);
		
		\node [draw, ultra thick, circle, minimum width=10pt, color=gray] (c2) at ([xshift=80]s2.east) {};
		\draw [->, ultra thick, color=black!20] (s2.east) --(c2.west) coordinate(p4);			
		\draw [ultra thick, color=gray] ([xshift=1]c2.west) -- ([xshift=-1]c2.east);
		\draw [ultra thick, color=gray] ([yshift=1]c2.south) -- ([yshift=-1]c2.north);
		
		\node [ASTYLE, fill=rfcolor!30] (att1) at ([xshift=25, yshift=-20]s1.east) {$A_{dr}$};	
		\node [ASTYLE, fill=dfcolor!30] (att2) at ([xshift=25, yshift=20]s2.east) {$A_{rd}$};
		
		\draw [->, ultra thick, color=black!20] ([xshift=1]s1.east) to[bend right=20] (att1.west);
		\draw [->, ultra thick, color=black!20] ([xshift=1]s2.east) to[bend left=20] (att2.west);
		\draw [->, ultra thick, color=black!20] ([xshift=1]s2.east) to [bend left=20] (att1.south west);
		\draw [->, ultra thick, color=black!20] ([xshift=1]s1.east) to[bend right=20] (att2.north west);
		\node [draw, ultra thick,circle, minimum width=10pt, color=gray] (g1) at ([xshift=20]att1.east) {};
		\node [draw, ultra thick,circle, minimum width=10pt, color=gray] (g2) at ([xshift=20]att2.east) {};
		\node [draw, circle, minimum size=1pt, inner sep=0pt, line width=1pt, color=gray] at (g1){};
		\node [draw, circle, minimum size=1pt, inner sep=0pt, line width=1pt, color=gray] at (g2){};
		
		\node [] (g1t) at ([yshift=-12]g1.south) {$g_1$};
		\node [] (g2t) at ([yshift=12]g2.north) {$g_2$};
		\draw [->, ultra thick, color=black!20] (g1t.north) -- (g1.south);
		\draw [->, ultra thick, color=black!20] (g2t.south) -- (g2.north);
		\draw [->, ultra thick, color=black!20] (att1.east) -- (g1.west);
		\draw [->, ultra thick, color=black!20] (att2.east) -- (g2.west);
		\draw [->, ultra thick ,color=black!20] (g1.east) to [bend right=30] (c1.south);
		\draw [->, ultra thick, color=black!20] (g2.east) to [bend left=30] (c2.north);
		\coordinate (g12) (g1t.east) + (g2t.east);
		\node [scale=0.9] (eq1)at ([xshift=30]g12) {$g_1 + g_2 = 1$};
		\node [below=1pt of eq1, scale=0.9, xshift=11] {$g_1 = \hat{g}$};
		\draw [->, ultra thick, color=rfcolor!45] (c1.east) -- ++(1.2,0) coordinate(p5);
		\draw [->, ultra thick, color=dfcolor!45] (c2.east) -- ++(1.2,0) coordinate(p6);
		\node [] at ([xshift=8.]p5) {$\mathrm{rf}_i$};
		\node [] at ([xshift=8.]p6) {$\mathrm{df}_i$};
		\node [] at ([xshift=13, yshift=-7]s1.east) {$\tilde{\mathrm{rb}}_i$};
		\node [] at ([xshift=13, yshift=7]s2.east) {$\tilde{\mathrm{db}}_i$};
		
		\networkLayer{1}{0.2}{4}{0}{1}{color=rdcolor!80}{}{$\tilde{\mathrm{rb}}_i$}
		\networkLayer{1}{0.2}{4}{0}{-1}{color=ddcolor!40}{$\tilde{\mathrm{db}}_i$}{}
		
		\networkLayer{1}{0.2}{6}{0}{1.4}{color=color3!40}{}{}
		\networkLayer{1}{0.2}{6}{0}{-1+.8}{color=color4!40}{}{}
		\networkLayer{1}{0.2}{6}{0}{-1-.8}{color=color9!40}{}{}
		\draw [->, ultra thick, color=blue!40] (4, -.8) -- ++(1.5, .8);
		\node [rotate=28] at (4.8, -0.1) {1x1 conv};
		\draw [->, ultra thick, color=blue!40] (4, -.8) -- ++(1.5, -.8);
		\node [rotate=-28] at (4.8, -1.4) {1x1 conv};
		\draw [->, ultra thick, color=blue!40] (4, 1.1) -- ++(1.5, .5);
		\node [rotate=20] at (4.85, 1.6) {1x1 conv};
		\node [scale=0.8] at (4, -2.2) {$C\times H\times W$};
		\node [scale=.8] at (4, 0.4) {$C\times H\times W$};
		\draw [->, ultra thick, color=blue!40] (6, -1.6) -- ++(1.8, 0) --++(0, .6);
		\draw [->, ultra thick, color=blue!40] (6, .1) --++(1.8, 0)--++(0, -.7);
		\node [scale=.8] at (7, -1.9) {$C_1\times (HW)$};
		\node [scale=.8] at (6.5, -1.4) {$W_k$};
		\node [scale=.8] at (7, -.2) {$(HW)\times C_1$};
		\node [scale=.8] at (6.5, 0.35) {$W_q^T$};
		\node [draw, ultra thick,circle, minimum width=10pt, color=gray] (m1)  at (7.8, -0.8) {};
		\draw [ultra thick, color=gray] ([xshift=1]m1.north west) -- ([xshift=-1]m1.south east);
		\draw [ultra thick, color=gray] ([xshift=1]m1.south west) -- ([xshift=-1]m1.north east);
		\draw [->, ultra thick, color=blue!40] (m1.east) -- ++(1.4, 0) coordinate(pm1);	
		\node [] at ([xshift=18, yshift=5] m1.east) {softmax};
		\node [draw, rectangle, xshift=20, minimum width=40, minimum height=40, ball color=blue!40,shading=ball] (w1) at (pm1.east) {};
		\node [scale=.8] at ([yshift=-9]w1.south) {$(HW)\times (HW)$};
		\node [scale=.8] at ([yshift=5]w1.north) {$W_a$};
		\node [scale=.8] at (7, 1.3) {$C\times (HW)$};
		\node [scale=.8] at (6.5, 1.8) {$W_v$};
		\node [draw, ultra thick,circle, minimum width=10pt, color=gray] (m2)  at ([xshift=20]w1.east) {};	
		\draw [->, ultra thick, color=blue!40] (w1.east) --(m2.west);
		\draw [->, ultra thick, color=blue!40] (6, 1.6) --++(5.5, 0) --++(0, -2.2);
		\draw [ultra thick, color=gray] ([xshift=1]m2.north west) -- ([xshift=-1]m2.south east);
		\draw [ultra thick, color=gray] ([xshift=1]m2.south west) -- ([xshift=-1]m2.north east);	
		\draw [->, ultra thick, color=blue!40] (m2.east) -- ++(1.5,0);
		\node [scale=.8]at ([xshift=18, yshift=-8] m2.east) {$C\times H\times W$};
		\node [] at ([xshift=50]m2.east){$f_{dr}$};
		\node [] at ([yshift=-10]b1.south) (a) {(a)};
		\node [] at ([xshift=250]a) {(b)};	
		\end{tikzpicture}
		
	}
	\caption{Illustration of GMA module. (a) shows the constructure of GMA module, and (b) represents the operation $A_{dr}$. The operation $A_{rd}$ is symmetrical to the $A_{dr}$ (exchange the position of $\tilde{\mathrm{rb}}_i$ and $\tilde{\mathrm{db}}_i$). For conciseness, we just show the $A_{dr}$. And the $\hat{g}$ refers to the prediction score of the depth potentiality. }
	\label{fig:gma}
\end{figure*}
Most previous works \cite{han2017cnns,zhu2019pdnet,chen2018progressively,zhao2019contrast,Piao_2019_ICCV} generally integrate the multi-modal features from RGB and corresponding depth information indiscriminately. However, as mentioned before, there exist some contaminations when depth maps are unreliable. To address this issue, Fan \etal \cite{fan2019D3Net} proposed a depth depurator unit to switch the RGB path and RGB-D path in a mechanical and unsupervised way. Different from the work \cite{fan2019D3Net}, our proposed network can explicitly model the confidence response of the depth map and control the fusion process in a soft manner rather than directly discard the low-quality depth map.\\
\indent Since we do not hold any labels for depth map quality assessment, we model the depth potentiality perception as a saliency-oriented prediction task, that is, we train a model to automatically learn the relationship between the binary depth map and the corresponding saliency mask. The above modeling approach is based on the observation that if the binary depth map segmented by a threshold is close to the ground truth, the depth map is highly reliable, so a higher confidence response should be assigned to this depth input. Specifically, we first apply Otsu \cite{otsu1979threshold} to binarize the depth map $I$ into a binary depth map $\tilde{I}$, which describes the potentiality of the depth map from the  saliency perspective. Then, we design a measurement to quantitatively evaluate the degree of correlation between the binary depth map and the ground truth. IoU (intersection over union) is adopted to measure the accuracy between the binary map $\tilde{I}$ and the ground truth $G$, which can be formulated as:
\begin{equation}
 D_{\mathrm{iou}} = \frac{|\tilde{I}\cap G|}{|\tilde{I}\cup G|}, \label{eq:m1}
\end{equation}
where $|\cdot|$ denotes the area.
However, in some cases, the coarse binary depth map will contain the background, which causes the $D_{\mathrm{iou}}$ tending to small even if the final saliency map is very closed to the ground truth. Hence, we define another metric to relax the strong constraint of IoU, which is defined as:
\begin{equation}
 D_{\mathrm{cov}} = \frac{|\tilde{I}\cap G|}{|G|}. \label{eq:m2}
\end{equation}
This metric $D_{\mathrm{cov}}$ reflects the ratio of intersection area to the ground truth, which indicates that the binary depth map is expected to cover the salient object more complete.
Finally, inspired by the F-measure \cite{achanta2009frequency}, we combine these two metrics to measure the potentiality of depth map for SOD task, \ie,
\begin{equation}
  D(\tilde{I}, G) = \frac{(1 + \gamma)\cdot D_{\mathrm{iou}} \cdot D_{\mathrm{cov}}}{
  	D_{\mathrm{iou}} + 	\gamma \cdot D_{\mathrm{cov}}}, \label{eq:m3}
\end{equation}
where $\gamma$ is a weighting coefficient. Considering the noise and inaccuracy that may be caused by threshold segmentation, we put more emphasis on the completeness of covering regions when combining the IoU metric and COV metric, thus we set the $\gamma$ to $0.3$ to emphasis the $D_{\mathrm{cov}}$ over $D_{\mathrm{iou}}$ following the setting in \cite{achanta2009frequency}.\\

\indent To learn the potentiality of the depth map, we provide $D({\tilde{I}, G})$ as the pseudo label $g$ to guide the training of the regression process. Specifically, the top layers features of two branches' backbone are concatenated after passing through GAP, and then two fully connected layers are applied to obtain the estimation $\hat{g}$. The $D(\tilde{I}, G)$ is only available in the training phase. As $\hat{g}$ reflects the potentiality confidence of depth map, we introduce it in the GMA module to prevent the contamination from unreliable depth map in the fusion process, which will be explained in the GMA module.

\subsection{Gated Multi-modality Attention Module}
Taken into account that there exist complementarity and inconsistency of the cross-modal RGB-D data, directly integrating the cross-modal information may induce negative results, such as contaminations from unreliable depth maps. Besides, the features of the single modality usually are affluent in spatial or channel aspect, but also include information redundancy.
To cope with these issues, we design a GMA module that exploits the attention mechanism to automatically select and strengthen important features for saliency detection, and incorporate the gate controller into the GMA module to prevent the contamination from the unreliable depth map. \\
\indent To reduce the redundancy of single-modal features and highlight the feature response on the salient regions, we apply spatial attention (see `$S$' in Fig. \ref{fig:gma}) to the input feature $\mathrm{rb}_i$ and $\mathrm{db}_i$, respectively. The process can be described as:
\begin{align}
&f = conv_1 (f_{in}), \\
&(W;B) = conv_2 (f), \\
&f_{out} = \delta(W \odot f + B),
\end{align}
where $f_{in}$ represents the input feature of the RGB branch or depth branch (\ie, $\mathrm{rb}_i$ or $\mathrm{db}_i$), $conv_i$ ($i=1,2$) refers to the convolution layer, $\odot$ denotes element-wise multiplication, $\delta$ is the ReLU activation function, and $f_{out}$ represents the modified RGB/depth feature (\ie, $\tilde{\mathrm{rb}_i}$ or $\tilde{\mathrm{db}_i}$).
The channels of modified feature $\tilde{\mathrm{rb}_i}$, $\tilde{\mathrm{db}_i}$ are unified into $256$ dimensions at each stage. Note that, the weights are not shared for the RGB and depth branches in our model.\\
\indent Further, inspired by the success of self-attention \cite{vaswani2017attention,wang2018non}, we design two symmetrical attention sub-modules to capture long-range dependencies from a cross-modal perspective. Taking $A_{dr}$ in Fig. \ref{fig:gma} as an example,
$A_{dr}$ exploits the depth information to generate a spatial weight for RGB feature $\tilde{\mathrm{rb}}_i$, as depth cues usually can provide helpful information (\eg, the coarse location of salient objects) for RGB branch.
Technically, we first apply $1\times 1$ convolution operation to project the $\tilde{\mathrm{db}}_i$ into $W_q \in \mathbb{R}^{C_1\times (HW)}$, $W_k \in \mathbb{R}^{C_1\times (HW)}$, and project the $\tilde{\mathrm{rb}}_i$ into $W_v \in \mathbb{R}^{C\times (HW)}$, where $C$, $H$, $W$ refer to the channel, height, width of the feature $W_v$, respectively, and $C_1$ is set to $1/8$ of $C$ for computation efficiency.
We compute the enhanced feature as follows:
\begin{align}
& W_a = softmax(W_q^T \otimes W_k), \\
& f_{dr} = W_v \otimes W_a,
\end{align}
where $softmax$ is applied in the column of $W_a$, $\otimes$ represents matrix multiplication.
The enhanced feature $f_{dr}$ is then reshaped into $C\times H\times W$.
The another sub-module $A_{rd}$ is symmetric to $A_{dr}$.
These two attention modules aim to capture the long-range dependencies from a cross-modal perspective, where $A_{dr}$ exploits the depth information to generate a spatial weight for the RGB feature, and $A_{rd}$ refines the depth feature by using the spatial weight generated from the RGB feature.
\\
\indent Finally, we introduce the gates $g_1$ and $g_2$ with the constraint of $g_1+g_2=1$ to control the interaction of the enhanced features and modified features, which can be formulated as
\begin{align}
&\mathrm{rf}_i = \tilde{\mathrm{rb}}_i + g_1 \cdot f_{dr}, \label{eq:rfi} \\
&\mathrm{df}_i = \tilde{\mathrm{db}}_i + g_2 \cdot f_{rd}.
\end{align}
where $\mathrm{rf}_i$ and $\mathrm{df}_i$ are the refined feature of the RGB and depth branches, respectively, which is used in the decoder stage.
In equation (\ref{eq:rfi}), we use the gate to control the interaction of the enhanced features and modified features. For the enhanced feature, the weight $g_1=\hat{g}$, where $\hat{g}$ is learned under the supervision of the pseudo label $g$. When $\hat{g}$ is closed to $1$, it means that the depth map is highly reliable, and more depth information will be introduced to RGB branch to reduce the background disturbances; When $\hat{g}$ is closed to 0, RGB branch will be the dominant branch and less depth information will be adopted, and the RGB information will play a more important role to prevent the contamination. Some feature visualization maps are shown in Fig. \ref{fig:att}. Taking the third image as an example, the quality of the depth map is poor intuitively, so the enhanced feature $f_{dr}$ is wrongly focused on the lower left areas. But with the constraint of the weight $g_1$, the encoder feature we get can effectively highlight the salient region and suppress the effect of enhanced feature $f_{dr}$. More details will be discussed in Section IV-E.

\subsection{Multi-level Feature Fusion}
Feature fusion plays a more critical role in RGB-D saliency detection due to cross-modal information, often directly affecting the performance. In order to obtain more comprehensive and discriminative fusion features, we consider two aspects for multi-level feature integration. First, the features at different scales contain different information, which can complement each other. Therefore, we use a multi-scale progressive fusion strategy to integrate the single-modal feature from coarse to fine. Second, for the multi-modality features, we exploit the designed GMA module to enhance the features separately instead of early fusing the RGB and depth features, which can reduce the interference of different modality. Finally, we aggregate the two modal features by using multi-modality feature fusion to obtain the saliency map.\\
\indent \textbf{Multi-scale Feature Fusion.} Low-level features can provide more detail information, such as boundary, texture, and spatial structure, but may be sensitive to the background noises. Contrarily, high-level features contain more semantic information, which is helpful to locate the salient object and suppress the noises. Different from previous works \cite{Qin_2019_CVPR,Liu2019PoolSal} generally fuse the low-level features and high-level features  by concatenation or summation operation, we adopt a more aggressive yet effective operation, \ie, multiplication. The multiplication operation can strengthen the response of salient objects, meanwhile suppress the background noises. Specifically, taking the fusion of higher level feature $\mathrm{rd}_5$ and lower level feature $\mathrm{rf}_4$ as an instance, the multi-scale feature fusion can be described as
\begin{align}
&f_1 = \delta(upsample({conv}_3(\mathrm{rd}_5)) \odot \mathrm{rf}_4), \label{eq:12} \\
&f_2 = \delta (conv_4 (\mathrm{rf}_4) \odot upsample(\mathrm{rd}_5)),\label{eq:13}  \\
&f_{F} = \delta(conv_5 ([f_1, f_2])),
\end{align}
where $upsample$ is the up-sampling operation via bilinear interpolation, and $[\cdot,\cdot]$ represents the concatenation operation.
The fusion result $f_{F}$ is exactly the higher level feature of the next fusion stage. \\
\indent\textbf{Multi-modality Feature Fusion.}
During the multi-modality feature fusion, we consider two issues:
(1) How to select the most useful and complementary information from the RGB and depth features. Thus, we learn the weight $\bm{\alpha}$ to balance the complementarity when combining the complementary information from two modal data. The weight $\bm{\alpha}$ is learned from the top layers features of two branches backbone, which demonstrates the channel importance of the multi-modality features.
(2) How to prevent the contamination caused by the unreliable depth map during fusing. Thus, the weight $\hat{g}$ is used to control the introduction ratio of the depth information. The weight $\hat{g}$ is learned under the supervision of the pseudo label g and reflects the potentiality confidence of depth map.

Specifically, to fuse the cross-modal features $\mathrm{rd}_2$ and $\mathrm{dd}_2$, we design a weighted channel attention mechanism to automatically select useful channels, which can be formulated as
\begin{align}
&f_3 = \bm{\alpha} \odot \mathrm{rd}_2 + \hat{g} \cdot (1 - \bm{\alpha}) \odot \mathrm{dd}_2, \label{eq:1}\\
&f_4 = \mathrm{rd}_2 \odot \mathrm{dd}_2, \label{eq:2}\\
&f_\mathrm{sal} = \delta(conv([f_3, f_4])),
\end{align}
where $\bm{\alpha} \in \mathbb{R}^{256}$ is the weight vector learned from RGB and depth information (see Fig. \ref{fig:main_plot}), $\hat{g}$ is the learned weight of the gate as mentioned before. The equation (\ref{eq:2}) reflects the common response for salient objects, while equation (\ref{eq:1}) combines the two modal features via channel selection ($\bm{\alpha}$) and gate mechanism ($\hat{g}$) for considering the complementarity and inconsistency.

\subsection{Loss Function}
For training the network, we consider the classification loss and regression loss to define the loss function, where the classification loss is used to constrain the saliency prediction, and the regression loss aims to model the depth potentiality response. \\
\indent \textbf{Classification Loss.}
In saliency detection, binary cross-entropy loss is commonly adopted to measure the relation between predicted saliency map and the ground truth, which can be defined as
\begin{equation}
\ell = -\frac{1}{H\times W} \sum_{i=1}^{H}\sum_{j=1}^{W}[G_{ij}\log(S_{ij})+ (1-G_{ij})\log(1 - S_{ij})] ,
\end{equation}
where $H$, $W$ refer to the height and width of the image respectively,
$G$ denotes the ground truth, and $S$ represents the predicted saliency map. To facilitate the optimization of the proposed network, we add auxiliary loss at four decoder stages. Specifically, a $3\times 3$ convolution layer is applied for each stage ($\mathrm{rd}_i$, $\mathrm{dd}_i$, $i=5,4,3,2$) to squeeze the channel of the output feature maps to 1. Then these maps are up-sampled to the same size as the ground truth via bilinear interpolation and sigmoid function is used to normalize the predicted values into $[0,1]$. Thus, the whole classification loss consists of two parts, \ie, the dominant loss corresponding to the output and the auxiliary loss of each sub-stage.
\begin{equation}
 \ell_{\mathrm{cls}} = \ell_{\mathrm{dom}} + \sum_{i=1}^{8} \lambda_{i}\ell_{\mathrm{aux}}^i ,
\end{equation}
where $\lambda_i$  denotes the weight of different loss, and $\ell_{\mathrm{dom}}$,
$\ell_{\mathrm{aux}}^i$ denote the dominant and auxiliary loss, respectively. The auxiliary loss branches only exist during the training stage. \\
\indent\textbf{Regression Loss.} To model the potentiality of depth map, the smooth L1 loss \cite{girshick2015fast} is used as the supervision signal. The smooth L1 loss  is defined as
\begin{equation}
 \ell_{\mathrm{reg}} = \left\lbrace 	 \
 \begin{aligned}
	 0.5 (g - \hat{g})^2,  &\quad \mathrm{if}\ |g - \hat{g}| < 1 \\
	 |g - \hat{g}| - 0.5, &\quad \mathrm{otherwise}
 \end{aligned}\right. ,
\end{equation}
where $g$ is the pseudo label as mentioned in the depth potentiality perception, and $\hat{g}$ denotes the estimation of the network as shown in Fig. \ref{fig:main_plot}. \\
\indent\textbf{Final Loss.} The final loss is the linear combination of the classification loss and regression loss,
\begin{equation}
	\ell_{\mathrm{final}} = \ell_{\mathrm{cls}} + \lambda \ell_{\mathrm{reg}},
\end{equation}
where $\lambda$ is weight of $\ell_{\mathrm{reg}}$, which is set to $1$ in our model.
The whole training process is conducted in an end-to-end way.


\section{Experiments}
\subsection{Datasets}
We evaluate the proposed method on $8$ public RGB-D SOD datasets with the corresponding pixel-wise ground-truth.

\begin{figure*}[!t]
	\centering
	\includegraphics[width=1\linewidth,height=0.5\linewidth]{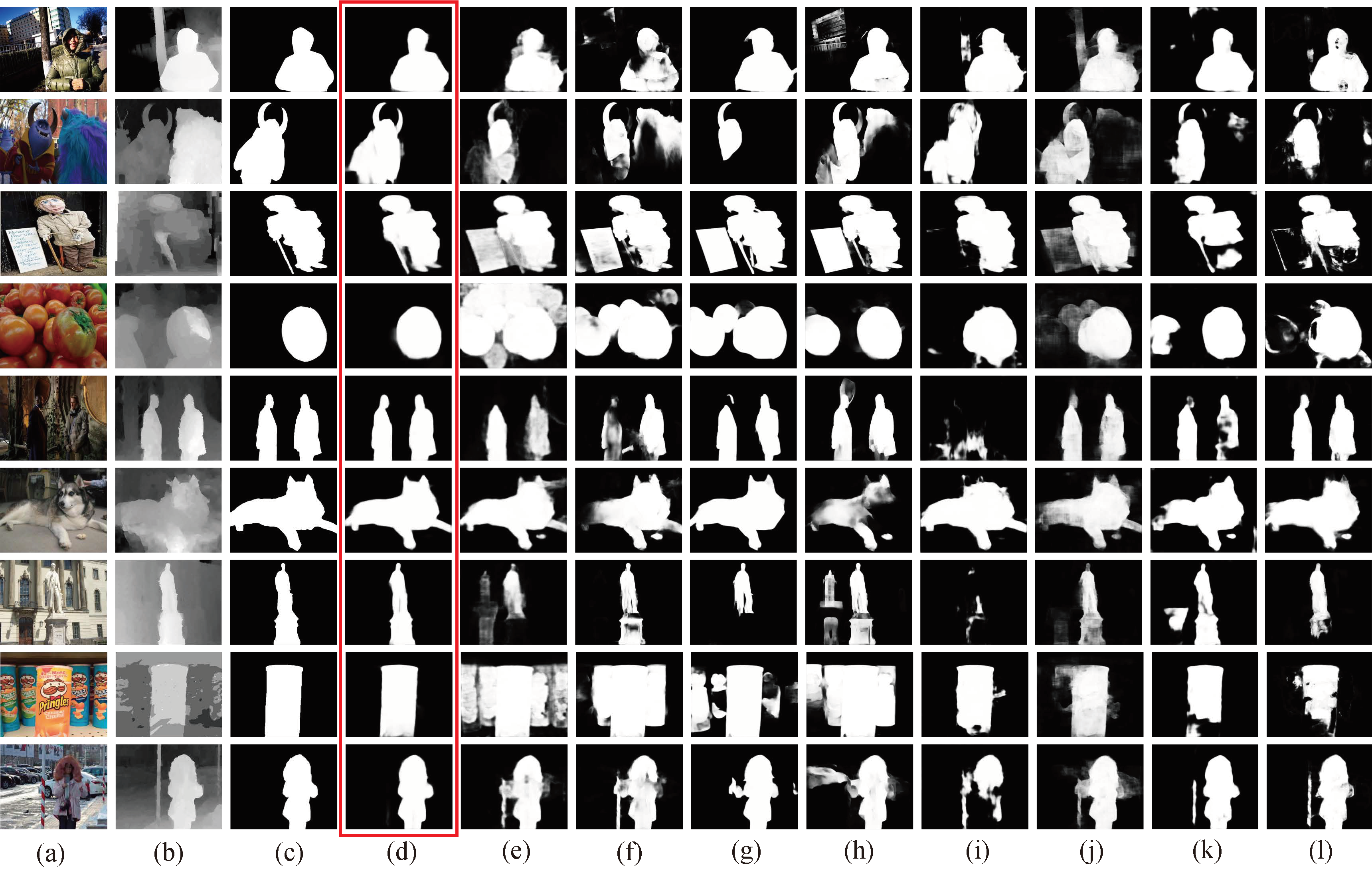}
	\caption{Qualitative comparison of the proposed approach with some state-of-the-art RGB and RGB-D SOD methods, in which our results are highlighted by a red box. (a) RGB image. (b) Depth map. (c) GT. (d) DPANet. (e) PiCAR. (f) PoolNet. (g) BASNet. (h) EGNet. (i) CPFP. (j) PDNet. (k) DMRA. (l) AF-Net.}
	\label{fig:vis}
\end{figure*}

\begin{figure*}[t]
	\centering
	\includegraphics[width=1\linewidth,height=0.4\linewidth]{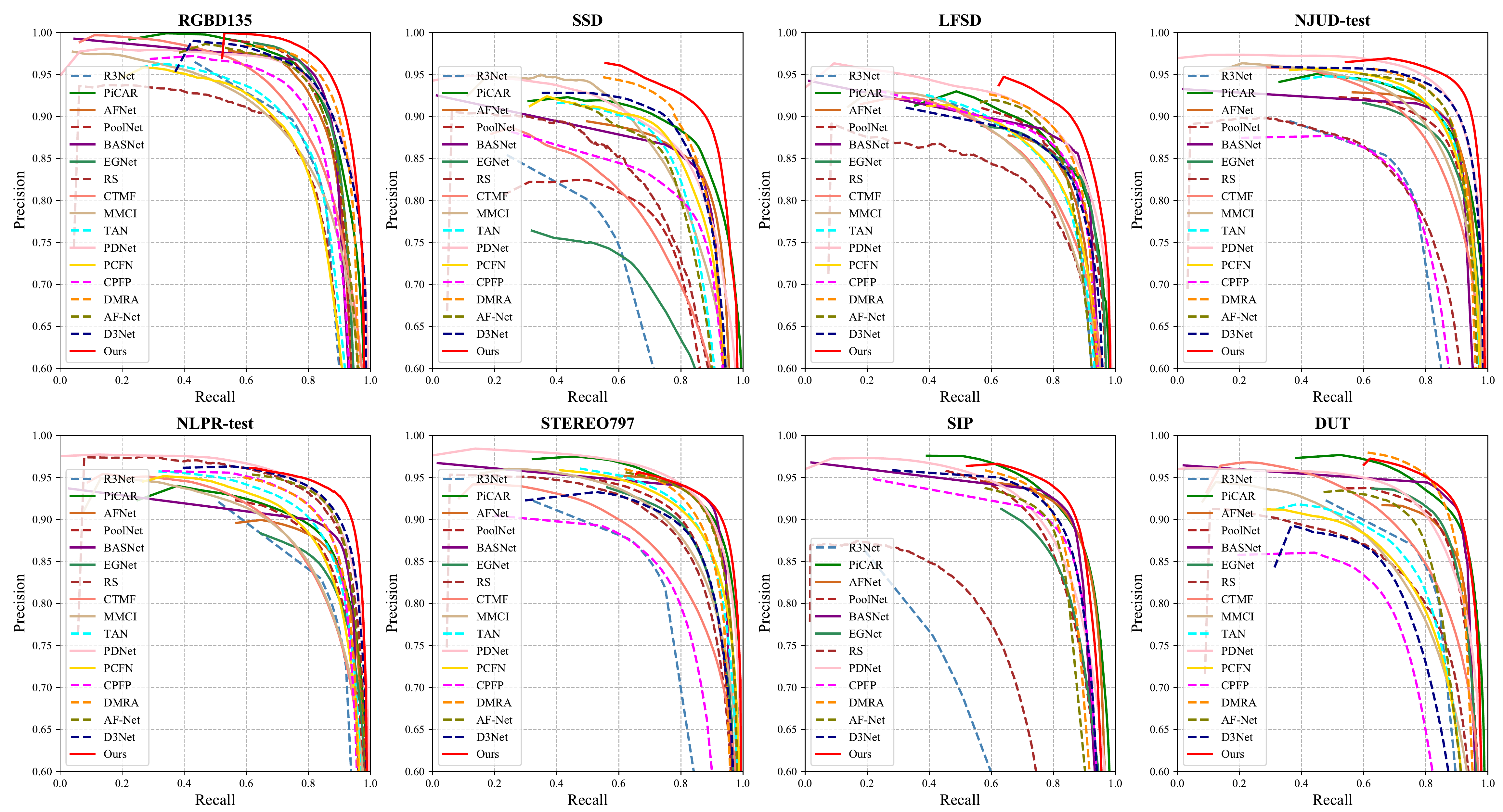}
	\caption{Illustration of PR curves on different datasets. The closer the PR curve is to $(1,1)$, the better performance of the method.}
	\label{fig:pr}
\end{figure*}

\textbf{NJUD} \cite{ju2014depth} consists of $2003$ RGB images and corresponding depth images with diverse objects and complex scenarios. The depth images are estimated from the stereo images.
\textbf{NLPR} \cite{peng2014rgbd} contains $1000$ RGB-D images captured by Kinect. Moreover, there exist multiple salient objects in an image of this dataset.
\textbf{STEREO797} \cite{niu2012leveraging} contains $797$ stereoscopic images collected from the Internet, where the depth maps are estimated from the stereo images.
\textbf{LFSD} \cite{li2014saliency} includes $100$ RGB-D images, in which the depth map is captured by Lytro light field camera.
\textbf{RGBD135} \cite{cheng2014depth} contains $135$ RGB-D images captured by Kinect simultaneously.
\textbf{SSD} \cite{zhu2017three} contains $80$ images picked up from three stereo movies, where the depth map is generated by depth estimation method.
\textbf{DUT} \cite{Piao_2019_ICCV} consists of $1200$ paired images containing more complex scenarios, such as multiple or transparent objects, \etc.
		This dataset is split into $800$ training data and $400$ testing data.
\textbf{SIP} \cite{fan2019D3Net} contains $929$ high-resolution person RGB-D images captured by Huawei Meta10.

\begin{table*}[t]
\renewcommand\arraystretch{1}
\caption{Performance comparison on $8$ public datasets. The best results on each dataset are highlighted in \textbf{boldface}.
From top to bottom: Our Method, CNN-based RGB-D SOD methods, and the latest RGB SOD methods. }
		\label{tab:pf1}
\vspace{-5 mm}
	\begin{center}
\resizebox*{1\textwidth}{!}{
\begin{tabular}{|c|c|c|c|c|}
				\hline
				\multirow{2}{*}{Method} & RGBD135 Dataset &SSD Dataset &LFSD Dataset &NJUD-test Dataset\\
				\cline{2-5} &\metrics&\metrics&\metrics&\metrics\\\hline\hline
			 DPANet (ours)&  \triplets(\textbf{0.933}, \textbf{0.919}, \textbf{0.023})&\triplets(\textbf{0.898}, \textbf{0.889}, \textbf{0.042})&\triplets(\textbf{0.891}, \textbf{0.864}, \textbf{0.072})&\triplets(\textbf{0.930}, \textbf{0.921}, \textbf{0.035})\\			
\hline
                $\text{D}^3$Net (TNNLS20)& \triplets(0.909, 0.898, 0.031) &
                \triplets(0.861, 0.857, 0.058) & \triplets(0.840, 0.825, 0.095)  & \triplets(0.909, 0.900, 0.047)  \\
				AF-Net (Arxiv19)&\triplets(0.904, 0.892, 0.033)&\triplets(0.828, 0.815, 0.077)&\triplets(0.857, 0.818, 0.091)&\triplets(0.900, 0.883, 0.053)\\
				DMRA (ICCV19)&\triplets({0.921}, {0.911}, {0.026})&\triplets({0.874}, 0.857, 0.055)&\triplets({0.865}, 0.831, {0.084})&\triplets(0.900, 0.880, {0.052})\\
				CPFP (CVPR19)&\triplets(0.882, 0.872, 0.038)&\triplets(0.801, 0.807, 0.082)&\triplets(0.850, 0.828, 0.088)&\triplets(0.799, 0.798, 0.079)\\
				PCFN (CVPR18)&\triplets(0.842, 0.843, 0.050)&\triplets(0.845, 0.843, 0.063)&\triplets(0.829, 0.800, 0.112)&\triplets(0.887, 0.877, 0.059)\\
				PDNet (ICME19)&\triplets(0.906, 0.896, 0.041)&\triplets(0.844, 0.841, 0.089)&\triplets({0.865}, {0.846}, 0.107)&\triplets({0.912}, {0.897}, 0.060)\\
				TAN (TIP19)&\triplets(0.853, 0.858, 0.046)&\triplets(0.835, 0.839, 0.063)&\triplets(0.827, 0.801, 0.111)&\triplets(0.888, 0.878, 0.060)\\
				MMCI (PR19)&\triplets(0.839, 0.848, 0.065)&\triplets(0.823, 0.813, 0.082)&\triplets(0.813, 0.787, 0.132)&\triplets(0.868, 0.859, 0.079)\\
				CTMF (TCyb18)&\triplets(0.865, 0.863, 0.055)&\triplets(0.755, 0.776, 0.100)&\triplets(0.815, 0.796, 0.120)&\triplets(0.857, 0.849, 0.085)\\
				RS (ICCV17)&\triplets(0.841, 0.824, 0.053)&\triplets(0.783, 0.750, 0.107)&\triplets(0.795, 0.759, 0.130)&\triplets(0.796, 0.741, 0.120)\\
			    \hline
				EGNet (ICCV19)&\triplets(0.913, 0.892, 0.033)&\triplets(0.704, 0.707, 0.135)&\triplets(0.845, 0.838, 0.087)&\triplets(0.867, 0.856, 0.070)\\
				BASNet (CVPR19)&\triplets(0.916, 0.894, 0.030)&\triplets(0.842, 0.851, 0.061)&\triplets(0.862, 0.834, {0.084})&\triplets(0.890, 0.878, 0.054)\\
				PoolNet (CVPR19)&\triplets(0.907, 0.885, 0.035)&\triplets(0.764, 0.749, 0.110)&\triplets(0.847, 0.830, 0.095)&\triplets(0.874, 0.860, 0.068)\\
				AFNet (CVPR19)&\triplets(0.897, 0.878, 0.035)&\triplets(0.847, 0.859, 0.058)&\triplets(0.841, 0.817, 0.094)&\triplets(0.890, 0.880, 0.055)\\
				PiCAR (CVPR18)&\triplets(0.907, 0.890, 0.036)&\triplets(0.864, {0.871}, {0.055})&\triplets(0.849, 0.834, 0.104)&\triplets(0.887, 0.882, 0.060)\\
				$\text{R}^3$Net (IJCAI18)&\triplets(0.857, 0.845, 0.045)&\triplets(0.711, 0.672, 0.144)&\triplets(0.843, 0.818, 0.089)&\triplets(0.805, 0.771, 0.105)\\\hline
			\end{tabular}}
		\end{center}
\end{table*}

\subsection{Evaluation Metrics}
To quantitatively evaluate the effectiveness of the proposed method, precision-recall (PR) curves, F-measure ($F_\beta$) score and curves, Mean Absolute Error (MAE), and S-measure ($S_m$) are adopted.
Thresholding the saliency map at a series of values, pairs of precision-recall value can be computed by comparing the binary saliency map with the ground truth.
The F-measure\footnote{We report the maximum F-measure in our results} is a comprehensive metric that takes both precision and recall into account, which is defined as:
\begin{equation}
 F_{\beta} = \frac{(1+{\beta}^2)\cdot{Precision}\cdot{Recall}}{{\beta}^2\cdot Precision + Recall}
\end{equation}
 where $\beta^2$ is set to $0.3$ to emphasize the precision over recall, as suggested by \cite{achanta2009frequency}.

 MAE is defined as the average pixel-wise absolute difference between the saliency map and the ground truth:
 \begin{equation}
 MAE = \frac{1}{H\times W}\sum_{y=1}^{H}\sum_{x=1}^{W}|S(x, y) - G(x, y)|
\end{equation}
where $S$ is the saliency map, $G$ denotes the ground truth, $H$ and $W$ are the height and width of the saliency map, respectively.

S-measure evaluates the structural similarity between the saliency map and the ground truth \cite{fan2017structure}, which is defined as:
 \begin{equation}
S_m = \alpha * S_o + (1-\alpha) * S_r
 \end{equation}
where $\alpha$ is set to $0.5$ to balance the object-aware structural similarity ($S_o$) and region-aware structural similarity ($S_r$).

\subsection{Implementation Details}
Following \cite{chen2018progressively}, we take $1400$ images from NJUD \cite{ju2014depth} and $650$ images from NLPR \cite{peng2014rgbd} as the training, and $100$ images from NJUD dataset and $50$ images from NLPR dataset as the validation set. Moreover, we unified the depth map with the necessary flips before training. To reduce the overfitting, we use multi-scale resizing and random horizontal flipping augmentation. During the inference stage, images are simply resized to $256\times 256$, and then fed into the network to obtain prediction without any other post-processing (\eg, CRF \cite{krahenbuhl2011efficient}) or pre-processing techniques (\eg, HHA \cite{gupta2014learning}). We use Pytorch to implement our model, and the ResNet-50 \cite{he2016deep} is used as our backbone. Mini-batch stochastic gradient descent (SGD) is used to optimize the network with the batch size of $32$, the momentum of $0.9$, and the weight decay of $5e\text{-}4$. The loss weights are set as $\{\lambda_1,\lambda_2, \cdots,\lambda_8\}=\{1.0,0.8,0.6,0.4,1.0,0.8,0.6,0.4\}$. We use the warm-up and linear decay strategies with the maximum learning rate $5e\text{-}3$ for the backbone and $0.05$ for other parts and stop training after $30$ epochs.
\subsection{Comparison with the State-of-the-arts}

We compare the proposed model with $16$ state-of-the-art methods, including $10$ RGB-D saliency models (\ie, $\text{D}^3$Net \cite{fan2019D3Net}, AF-Net \cite{wang2019adaptive}, DMRA \cite{Piao_2019_ICCV}, CPFP \cite{zhao2019contrast}, PCFN \cite{chen2018progressively}, PDNet \cite{zhu2019pdnet}, TAN \cite{chen2019three}, MMCI \cite{chen2019multi}, CTMF \cite{han2017cnns}, and RS \cite{shigematsu2017learning}) and $6$ latest RGB saliency models (\ie, EGNet \cite{zhao2019EGNet}, BASNet \cite{Qin_2019_CVPR}, PoolNet \cite{Liu2019PoolSal}, AFNet \cite{feng2019attentive}, PiCAR \cite{liu2018picanet}, and $\text{R}^3$Net \cite{deng2018r3net}).
For fair comparisons, we use the released code and default parameters to reproduce the saliency maps or the saliency maps provided by the authors.
\\ 

\begin{table*}[ht]
\renewcommand\arraystretch{1}
\caption{Continuation of Table \ref{tab:pf1}.}
			\label{tab:pf2}
\vspace{-5 mm}
	\begin{center}
		\resizebox*{1\textwidth}{!}{
\begin{tabular}{|c|c|c|c|c|}
				\hline
				\multirow{2}{*}{Method}& NLPR-test Dataset &STEREO797 Dataset &SIP Dataset &DUT Dataset\\
				\cline{2-5} & \metrics&\metrics&\metrics&\metrics\\\hline\hline
				DPANet (ours)&\triplets(\textbf{0.922}, \textbf{0.928}, \textbf{0.024})&\triplets(\textbf{0.915}, \textbf{0.911}, \textbf{0.041})&\triplets(\textbf{0.904}, \textbf{0.883}, \textbf{0.051})&\triplets(\textbf{0.916}, {0.899}, {0.048})\\				
\hline
                $\text{D}^3$Net (TNNLS20)& \triplets(0.907, 0.912, 0.030) & \triplets(0.869, 0.868, 0.058) & \triplets(0.881, 0.860, 0.063) & \triplets(0.797, 0.775, 0.097) \\
				AF-Net (Arxiv19)&\triplets(0.904, 0.903, {0.032})&\triplets(0.905, 0.893, 0.047)&\triplets(0.870, 0.844, 0.071)&\triplets(0.862, 0.831, 0.077)\\
				DMRA (ICCV19)&\triplets(0.887, 0.889, 0.034)&\triplets(0.895, 0.874, 0.052)&\triplets(0.883, 0.850, 0.063)&\triplets({0.913},  {0.880}, {0.052})\\
				CPFP (CVPR19)&\triplets(0.888, 0.888, 0.036)&\triplets(0.815, 0.803, 0.082)&\triplets(0.870, 0.850, 0.064)&\triplets(0.771, 0.760, 0.102)\\
				PCFN (CVPR18)&\triplets(0.864, 0.874, 0.044)&\triplets(0.884, 0.880, 0.061)&\fillcell&\triplets(0.809, 0.801, 0.100)\\
				PDNet (ICME19)&\triplets({0.905}, 0.902, 0.042)&\triplets(0.908, 0.896, 0.062)&\triplets(0.863, 0.843, 0.091)&\triplets(0.879, 0.859, 0.085)\\
				TAN (TIP19)&\triplets(0.877, 0.886, 0.041)&\triplets(0.886, 0.877, 0.059)&\fillcell&\triplets(0.824, 0.808, 0.093)\\
				MMCI (PR19)&\triplets(0.841, 0.856, 0.059)&\triplets(0.861, 0.856, 0.080)&\fillcell&\triplets(0.804, 0.791, 0.113)\\
				CTMF (TCyb18)&\triplets(0.841, 0.860, 0.056)&\triplets(0.827, 0.829, 0.102)&\fillcell&\triplets(0.842, 0.831, 0.097)\\
				RS (ICCV17)&\triplets(0.900, 0.864, 0.039)&\triplets(0.857, 0.804, 0.088)&\fillcell&\triplets(0.807, 0.797, 0.111)\\\hline
				EGNet (ICCV19)&\triplets(0.845, 0.863, 0.050)&\triplets(0.872, 0.853, 0.067)&\triplets(0.846, 0.825, 0.083)&\triplets(0.888, 0.867, 0.064)\\
				BASNet (CVPR19)&\triplets(0.882, 0.894, 0.035)&\triplets({0.914}, {0.900}, {0.041})&\triplets({0.894}, 0.872, {0.055})&\triplets({0.912}, \textbf{0.902}, \textbf{0.041})\\
				PoolNet (CVPR19)&\triplets(0.863, 0.873, 0.045)&\triplets(0.876, 0.854, 0.065)&\triplets(0.856, 0.836, 0.079)&\triplets(0.883, 0.864, 0.067)\\
				AFNet (CVPR19)&\triplets(0.865, 0.881, 0.042)&\triplets(0.905, 0.895, 0.045)&\triplets({0.891}, {0.876}, {0.055})&\triplets(0.880, 0.868, 0.065)\\
				PiCAR (CVPR18)&\triplets(0.872, 0.882, 0.048)&\triplets(0.906, {0.903}, 0.051)&\triplets(0.890, {0.878}, 0.060)&\triplets(0.903, {0.892}, 0.062)\\
				$\text{R}^3$Net (IJCAI18)&\triplets(0.832, 0.846, 0.049)&\triplets(0.811, 0.754, 0.107)&\triplets(0.641, 0.624, 0.158)&\triplets(0.841, 0.812, 0.079)\\\hline
			\end{tabular}}

		\end{center}
\end{table*}

\subsubsection{\textbf{Qualitative Evaluation}}

To further illustrate the advantages of the proposed method, we provide some visual examples of different methods. As shown in Fig. \ref{fig:vis}, our proposed network obtains a superior result with precise saliency location, clean background, complete structure, and sharp boundaries, and also can address various challenging scenarios, such as low contrast, complex scene, background disturbance, and multiple objects. To be specific,

(a) Our model achieves more complete structure and sharp boundaries in the results. For example, in the second image, the horns and body of the salient object cannot be completely detected by the comparison methods, such as PoolNet \cite{Liu2019PoolSal}, CPFP \cite{zhao2019contrast}, and the background regions (\eg, cartoon character on the right) are wrongly retained. Similarly, in the sixth image, most methods fail to detect the left front leg of the salient dog (\eg, AF-Net \cite{wang2019adaptive}, DMRA \cite{Piao_2019_ICCV}, EGNet \cite{zhao2019EGNet}, and BASNet \cite{Qin_2019_CVPR}), and the detected object boundaries are blurred and inaccurate (\eg, PDNet \cite{zhu2019pdnet} and PiCAR \cite{liu2018picanet}). By contrast, our method yields a more complete structure and sharp boundaries.

\begin{figure*}[!t]
    \centering
    \includegraphics[width=\textwidth]{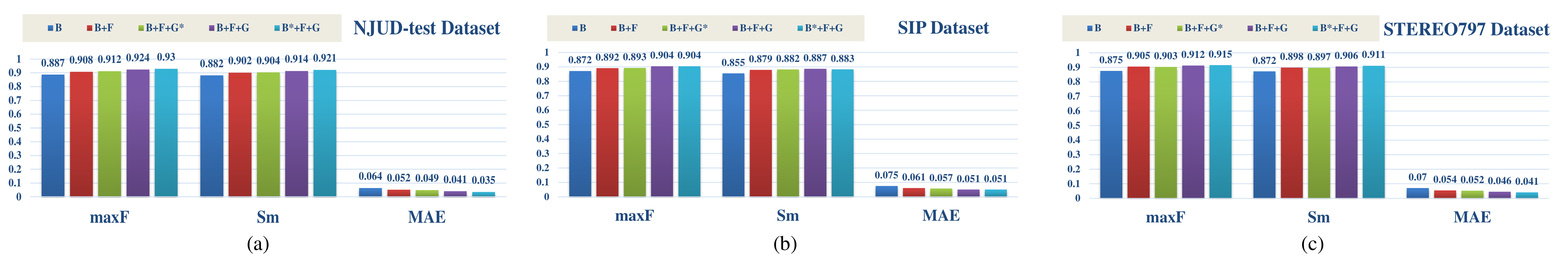}
    \caption{Ablation study of module verification on NJUD-test, SIP, and STEREO797 datasets.}
    \label{fig:ablation}
\end{figure*}

(b) Our model can address the complex and low contrast scenes. For example, in the fifth image, the low light makes the person in black on the left very close to the background, so many methods cannot accurately detect him, such as PoolNet \cite{Liu2019PoolSal}, CPFP \cite{zhao2019contrast}. By contrast, our model can detect those two persons with more complete structure and clean background. In the last image, the background is very complex, including a nearby indicator pole and multiple cars in the distance. Thus, the nearby indicator pole is detected as salient object by most methods, while our method can successfully suppress this region and obtain a better result with complete structure and clear boundaries.

(c) Our model can well handle the disturbance of similar appearances between the salient object and backgrounds. In the fourth image, there are many tomatoes in the image, but the slightly green tomato in the front is more prominent than others. Without the help of depth cues, all RGB SOD models fail to suppress these interferences, resulting in inaccurate detection results. Moreover, the existing RGB-D SOD methods also cannot address this challenging case very well. By contrast, our method shows a competitive advantage in terms of the completeness, sharpness, and accuracy. Analogously, our model can integrally highlight the sculpture and its pedestal in the seventh image, even the salient object has the similar color appearances with the background.

(d) Our model can produce more robust result even when confronting with the inaccurate or blurred depth information (\eg, the third and sixth images), which illustrates the power of the GMA module. In these challenging scenarios, it can be seen that the network is capable to utilize cross-modal complementary information and prevent the contamination from the unreliable depth map.

\subsubsection{\textbf{Quantitative Evaluation}}
For a more intuitive comparison of algorithm performance, we report the PR curves in Fig. \ref{fig:pr} and the quantitative metrics including the maximum F-measure, S-measure, and MAE score in Tables \ref{tab:pf1} and \ref{tab:pf2}. From the PR curves shown in Fig. \ref{fig:pr}, we can see that the proposed method achieves both higher precision and recall scores against other compared methods with an obvious margin on the eight datasets. As reported in Tables \ref{tab:pf1} and \ref{tab:pf2}, our method outperforms all the compared methods in terms of all the measurements, except that the MAE score on the DUT dataset achieves the second-best performance. It is worth mentioning that the performance gain is significant against the compared methods. For example, compared with the latest RGB-D SOD method \emph{DMRA}, our algorithm still achieves competitive performance in the case of a small number of training samples (\emph{i.e.}, the training samples do not include $800$ training data of the DUT dataset). On the SSD dataset, compared with the \emph{DMRA} method, the percentage gain of our \textit{DPANet} achieves $2.7\%$ in terms of maximum F-measure, $3.7\%$ in terms of S-measure, and $23.6\%$ in terms of MAE score. Compared with the \textbf{\emph{second best}} method in different datasets, our method still achieves obvious performance gain. For example, on the NJUD-test dataset, the percentage gain reaches $2.0\%$ for F-measure, $2.3\%$ for S-measure, and $25.5\%$ for MAE score. On the NLPR-test dataset, the \textbf{\emph{minimum percentage gain}} reaches $1.7\%$ for F-measure, $1.8\%$ for S-measure, and $20.0\%$ for MAE score.
Thus, all the quantitative measures demonstrate the effectiveness of the proposed model. \\

\subsubsection{\textbf{Run-time Comparison}}
We provide the run-time comparison results in the Table \ref{tab:speed}. From it, we can see that our model is faster than most of these deep learning based RGB-D SOD methods, which demonstrates the efficiency of the proposed \OURNET.
\begin{table}[!t]
    \centering
    \normalsize
    \caption{Comparisons of inference time of different deep learning based RGB-D SOD methods.}
    \begin{tabular}{|c|c|c|c|c|c|}
    \hline
         & CTMF & MMCI & TAN & PDNet & PCFN   \\
    \hline
     Time (s) & 0.63 & 0.05 & 0.07 & 0.07 & 0.06  \\
     \hline
         & CPFP & AF-Net & DMRA & $\text{D}^3$Net & Ours  \\
     \hline
     Time (s) & 0.17 & 0.03 & 0.06 & 0.05 & 0.03 \\
     \hline
    \end{tabular}
    \label{tab:speed}
\end{table}

\begin{figure*}[!t]
	\centering
	\includegraphics[width=1\textwidth]{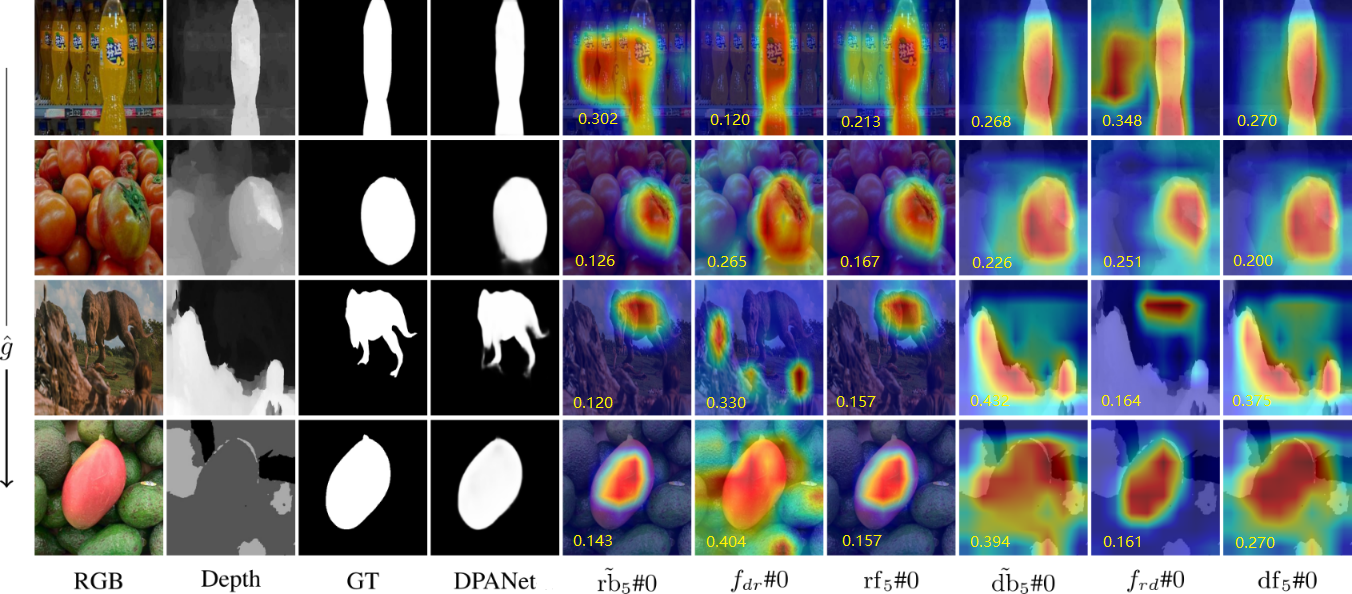}
	\caption {Visualization of the GMA module. ``\#$0$'' refers to the first channel of the features. The number at the bottom left of each image is the MAE value of the corresponding feature map.
	}
	\label{fig:att}
\end{figure*}

\begin{table*}[!t]
\centering
\normalsize
\renewcommand\arraystretch{1.1}
	\caption{Ablation studies on NJUD-test, SIP, and STEREO797 datasets. }
	\centering
\setlength{\tabcolsep}{3mm}
{
		\begin{tabular}{|c|c|c|c|}
	 	    \hline
			\multirow{2}{*}{}& NJUD-test Dataset &SIP Dataset &STEREO797 Dataset \\
			\cline{2-4} & \metrics&\metrics&\metrics\\\hline\hline
		    DPANet & \triplets({0.930}, {0.921}, {0.035}) & \triplets({0.904}, {0.883}, {0.051}) &\triplets({0.915}, {0.911}, {0.041}) \\
			\hline
			concatenation & \triplets(0.919, 0.914, 0.039) &\triplets(0.904, 0.876, 0.056) &\triplets(0.912, 0.905, 0.044) \\
			summation & \triplets(0.923, 0.915, 0.038) &\triplets(0.906, 0.881, 0.054) &\triplets(0.910, 0.904, 0.045) \\
			\hline
			hard manner & \triplets(0.908, 0.902, 0.047) & \triplets(0.893, 0.868, 0.064) & \triplets(0.905, 0.899, 0.050) \\
			\hline
			w/o depth & \triplets(0.908, 0.903, 0.043) &\triplets(0.864, 0.837, 0.074) &\triplets(0.913, 0.908, 0.042) \\
			\hline
		\end{tabular}}
	\label{tab:ablation}
\end{table*}

\subsection{Ablation Study}
\subsubsection{\textbf{Module Verification}} In this section, we conduct the ablation study to demonstrate the effectiveness of each key components designed in the proposed network on three different datasets: NJUD-test, SIP, and STEREO797, and the results are shown in Fig. \ref{fig:ablation}. We choose the network that removes the GMA module and regression loss, replaces the multi-scale feature fusion module with concatenation, and replaces the multi-modality feature fusion with multiplication and summation as the baseline (denoted as `B'). From Fig. \ref{fig:ablation}, compared the ``B'' with the ``B+F'', the multi-scale feature fusion (denoted as `F') improves the baseline about $2\sim 3$ points in terms of the maximum F-measure. After adding the GMA module without regression loss constrain (denoted as `$\text{G}^*$'), the F-measure rises up to $0.912$ on NJUD-test dataset, which is comparable with the state-of-the-art results. Furthermore, the performance is significantly enhanced after adding regression loss (B+F+G), which achieves the percentage gain of $16.3\%$ on NJUD-test compared with the (B+F+G*) case in terms of MAE score. Finally, adding the gate module in the multi-modality feature fusion (\ie, B*+F+G), our method yields the best performance result with the percentage gain of $4.8\%$ in terms of F-measure and $44.4\%$ in terms of MAE compared with the original baseline on NJUD-test dataset. The experiments on another two datasets, \ie, SIP, and STEREO797 also demonstrate the effectiveness of the designed components.

To better understand the attention mechanism designed in the GMA module, we visualize some feature maps and the corresponding heat-maps in Fig. \ref{fig:att}.
Taking the fourth GMA module as an example, as expected, the GMA module should learn the cross-modal complementarity from a cross-modal perspective and prevent the contamination from the unreliable depth map. Recall that the output of the fourth GMA module is denoted as $\mathrm{rf}_5= \tilde{\mathrm{rb}}_5 + g_1 \cdot f_{dr}$, $\mathrm{df}_5 = \tilde{\mathrm{db}}_5 + g_2 \cdot f_{rd}$, where $g_1 = \hat{g}$, and $g_1 + g_2 = 1$. As Fig.\ \ref{fig:att} shows,
\begin{itemize}
	\item when the depth map is reliable, but the RGB image encounters interference from the similar background (\eg, the first and second images), the features of the RGB branch ($\tilde{\mathrm{rb}}_5$\#$0$) usually fail to well focus on the area of salient object. By contrast, the features of depth map ($\tilde{\mathrm{db}}_5$\#$0$) can provide complementary information to enhance the feature maps ($f_{dr}$\#$0$) by suppressing the background noises. Further, the features of the RGB branch ($\tilde{\mathrm{rb}}_5$\#$0$) and the enhanced features ($f_{dr}$\#$0$) are combined with the weight of $g_1$ to obtain the features ($\mathrm{rf}_5$\#$0$) of the decoder stage.
	
\item  when the depth map tends to unreliable (\eg, the third and fourth images), we can see that the imperfect depth information almost has no impact on the features of the RGB branch thanks to the gate controller $g_2$ (compare $\tilde{\mathrm{rb}}_5$\#$0$ and $\mathrm{rf}_5$\#$0$), and introducing the RGB information to the depth branch leads the features of depth branch more convergent (from $\tilde{\mathrm{db}}_5$\#$0$ to $\mathrm{df}_5$\#$0$).
\end{itemize}

In a summary, the designed GMA module can learn the complementarity from a cross-modal perspective and prevent the contamination from the unreliable depth map. The output features of the GMA module will be aggregated progressively in the decoder stage via multi-scale feature fusion separately. Finally, to obtain the saliency map, the features of the RGB and depth branches are aggregated via the multi-modality feature fusion. All these ablation studies demonstrate the effectiveness of main components in our network, including the GMA module, regression loss, and multi-level feature fusion.

\subsubsection{\textbf{Fusion Method Verification}} For fusing the low-level features and high-level features, we adopt a more aggressive yet effective operation to replace the concatenation or summation operation, i.e., multiplication. The main reason is that the multiplication operation can strengthen the response of salient objects, meanwhile suppress the background noises. Some comparisons of the different fusion methods are shown in the Table \ref{tab:ablation}, replacing the multiplication in equations (\ref{eq:12}) and (\ref{eq:13}) with the concatenation or summation operation. From it, we can see that using multiplication is more effective than concatenation or summation. These experiments verified the hypothesis that multiplication can strengthen the response of the salient object and suppress the background noise.

\subsubsection{\textbf{Depth-related Information Verification}} In this subsection, we conduct the ablation studies of the depth-related information.

First, we investigate the difference between the soft manner and hard manner in the learning stage of the depth potentiality. The soft manner is the default setting that we use the regression as the supervision. The hard manner changes the regression as a classification problem, \ie, the pseudo label $g$ is binarized by a fixed threshold $0.5$, and binary cross entropy loss is adopted as the supervision signal, which forces the predict depth potentiality $\hat{g}$ tending to $0$ or $1$ rather than a smooth contiguous value. From Table \ref{tab:ablation}, we can see that the soft manner (\OURNET) is better than the hard manner (row `hard manner') in a holistic view. In addition, the soft manner is not affected by the threshold setting and is more robust.

Second, in order to verify the importance of depth information in RGB-D SOD, we report the experimental result of using only RGB image, which is denoted as `w/o depth' in Table \ref{tab:ablation}. As shown in Table \ref{tab:ablation}, with the help of the depth map, the final saliency performance is obviously improved. For example, on the NJUD-test dataset, the F-measure is improved from $0.908$ to $0.930$ with the percentage gain of $2.4\%$, and the MAE score is improved from $0.043$ to $0.035$ with the percentage gain of $18.6\%$.

Third, to further evaluate the depth potentiality, we calculate the mean value of the depth potentiality (DP) score using equations (\ref{eq:m1})-(\ref{eq:m3}) on $8$ datasets, which is shown in the Table \ref{tab:dp}. In Table \ref{tab:dp}, we also report the minimum percentage gain of the F-measure, which is denoted as `minPG-F'. For the dataset with the worst depth map quality (\ie, NLPR-test dataset with the DP score of $0.481$), the minimum gain of our \OURNET\ reaches $1.7\%$. In general, our \OURNET\ can well handle the poor depth perception potential as well as the good.


\begin{table}[!t]
    \centering
    \normalsize
    \caption{Depth potentiality (DP) scores on different datasets}
    \setlength{\tabcolsep}{1.2mm}
{
    \begin{tabular}{|c|c|c|c|c|}
        \hline
         &  RGBD135 & SSD & LFSD & NJUD-test \\
         \hline
         DP score & $0.495$ & $0.675$ & $0.755$ & $0.657$ \\
         \hline
         minPG-F & $1.3 \%$ & $2.7\%$ & $1.9\%$ & $2.0\%$ \\
         \hline
         & NLPR-test & STEREO797 & SIP & DUT\\
         \hline
         DP score & $0.481$ & $0.635$ & $0.683$ & $0.650$ \\
         \hline
         minPG-F & $1.7\%$ & $0.1\%$ & $1.1\%$ & $0.3\%$ \\
         \hline
    \end{tabular}}
    \label{tab:dp}
\end{table}

\subsection{Failure Cases}
\begin{figure}[!t]
    \centering
    \resizebox{\columnwidth}{!}{
    \begin{tikzpicture}[xscale=1, yscale=-1]
    \def\w{0.2\columnwidth}
    \def\h{0.06\textheight}
    \def\W{0.21\columnwidth}
    \def\H{0.064\textheight}
    \node [inner sep=0] at(0, 0) {\includegraphics[width=\w,
    height=\h]{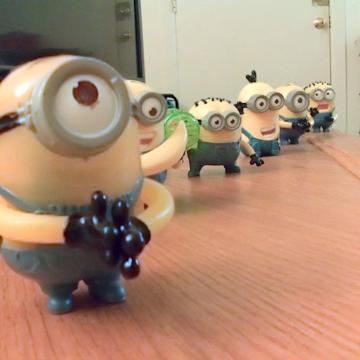}};
    \node [inner sep=0] at(\W, 0) {\includegraphics[width=\w, height=\h]{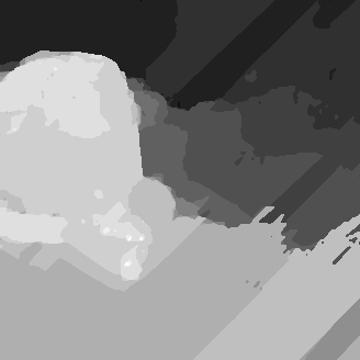}};
    \node [inner sep=0] at(\W*2, 0) {\includegraphics[width=\w, height=\h]{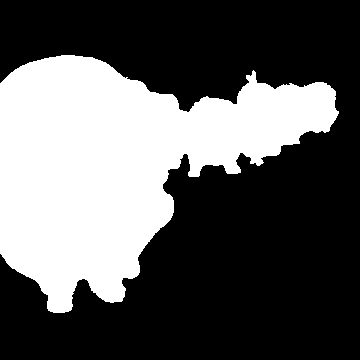}};
    \node [inner sep=0] at(\W*3, 0) {\includegraphics[width=\w, height=\h]{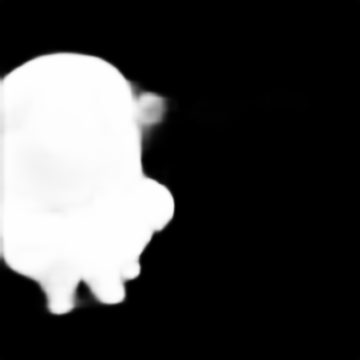}};
    \node [inner sep=0] at(\W*4, 0) {\includegraphics[width=\w, height=\h]{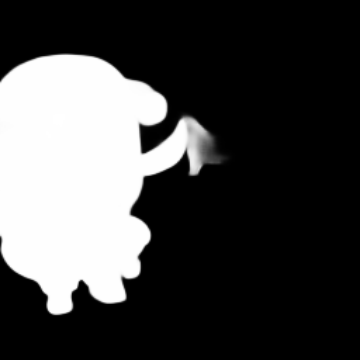}};
    \node [inner sep=0] at(\W*5, 0) {\includegraphics[width=\w, height=\h]{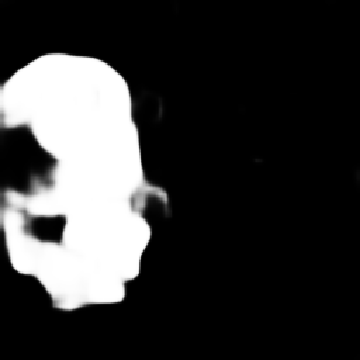}};
     \node [inner sep=0] at(0, \H) {\includegraphics[width=\w,
    height=\h]{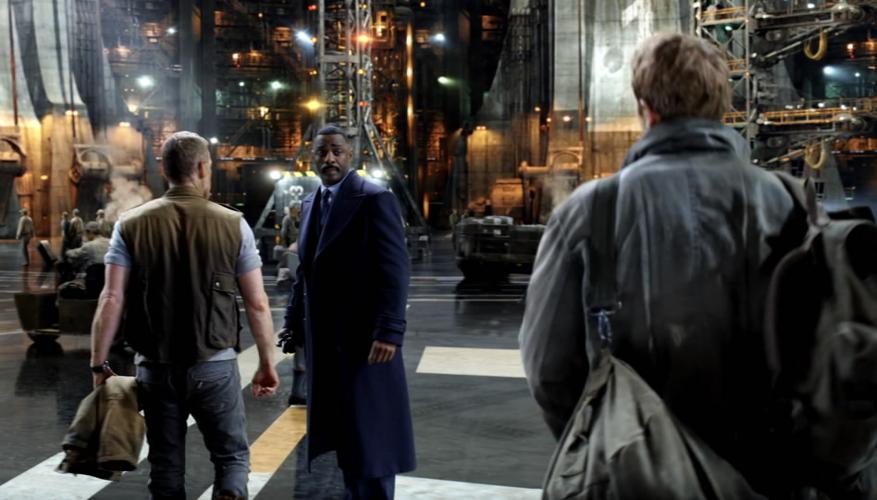}};
    \node [inner sep=0] at(\W, \H) {\includegraphics[width=\w, height=\h]{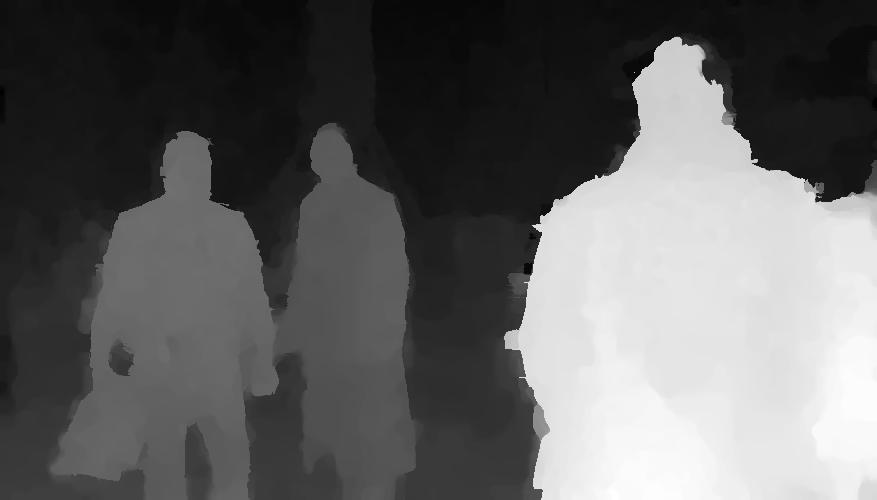}};
    \node [inner sep=0] at(\W*2, \H) {\includegraphics[width=\w, height=\h]{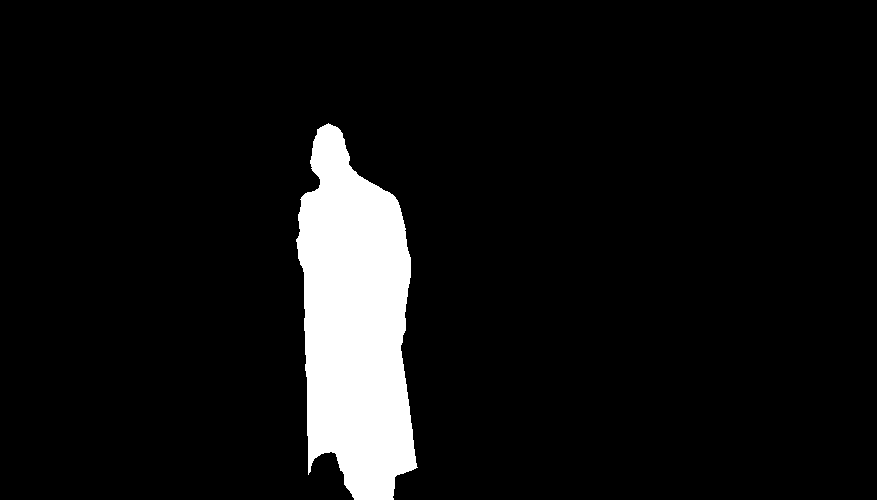}};
    \node [inner sep=0] at(\W*3, \H) {\includegraphics[width=\w, height=\h]{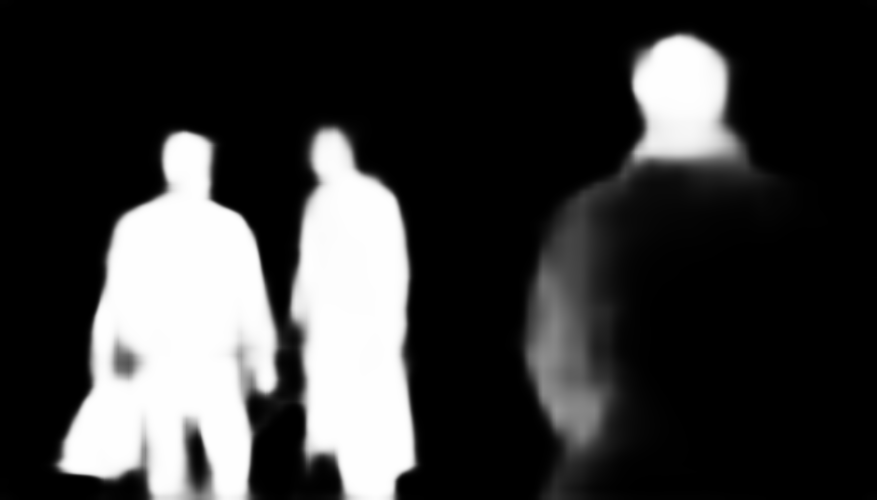}};
    \node [inner sep=0] at(\W*4, \H) {\includegraphics[width=\w, height=\h]{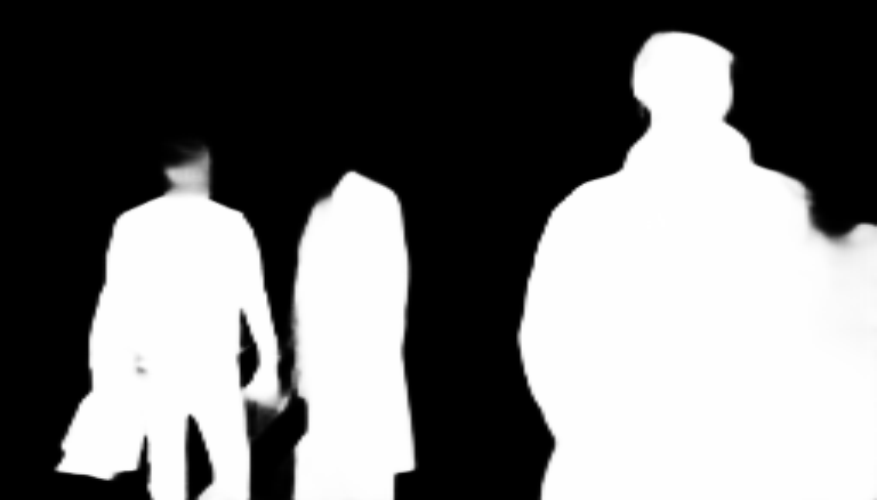}};
    \node [inner sep=0] at(\W*5, \H) {\includegraphics[width=\w, height=\h]{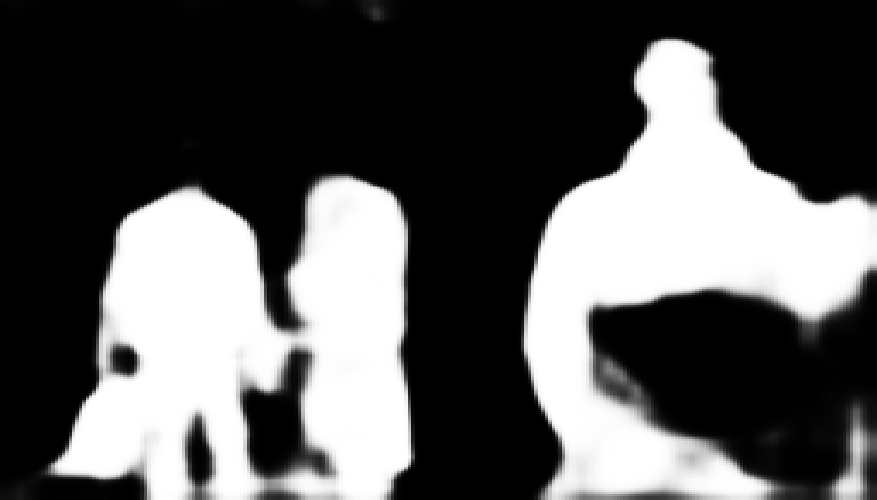}};
     \node [inner sep=0] at(0, \H*2) {\includegraphics[width=\w,
    height=\h]{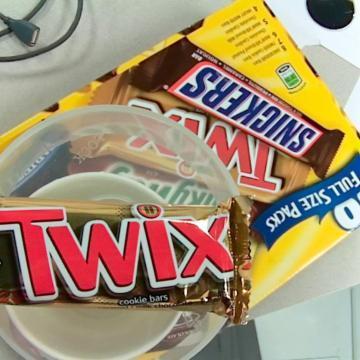}};
    \node [inner sep=0] at(\W, \H*2) {\includegraphics[width=\w, height=\h]{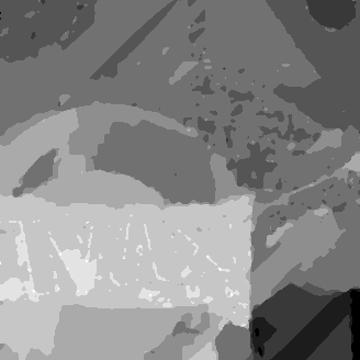}};
    \node [inner sep=0] at(\W*2, \H*2) {\includegraphics[width=\w, height=\h]{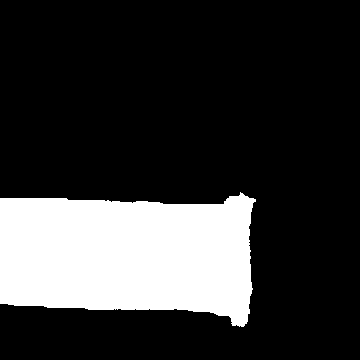}};
    \node [inner sep=0] at(\W*3, \H*2) {\includegraphics[width=\w, height=\h]{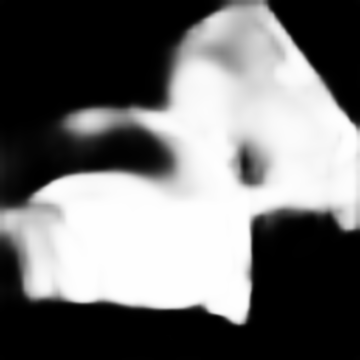}};
    \node [inner sep=0] at(\W*4, \H*2) {\includegraphics[width=\w, height=\h]{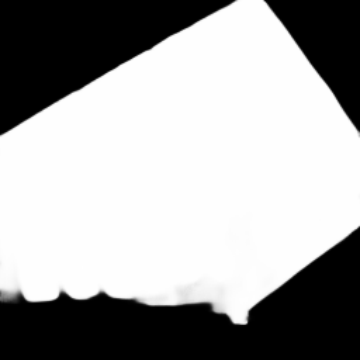}};
    \node [inner sep=0] at(\W*5, \H*2) {\includegraphics[width=\w, height=\h]{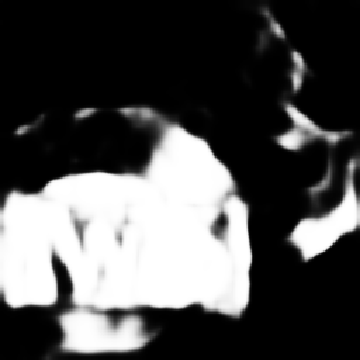}};

    \node [inner sep=0] at(0, \H*2.65) {\small RGB};
    \node [inner sep=0] at(\W, \H*2.65) {\small Depth};
    \node [inner sep=0] at(\W*2, \H*2.65) {\small GT};
    \node [inner sep=0] at(\W*3, \H*2.65) {\small Ours};
    \node [inner sep=0] at(\W*4, \H*2.65) {\small BASNet \cite{Qin_2019_CVPR}};
    \node [inner sep=0] at(\W*5, \H*2.65) {\small DMRA \cite{Piao_2019_ICCV}};
    \end{tikzpicture}
    }
    \caption{Failure examples.}
    \label{fig:failure}
\end{figure}
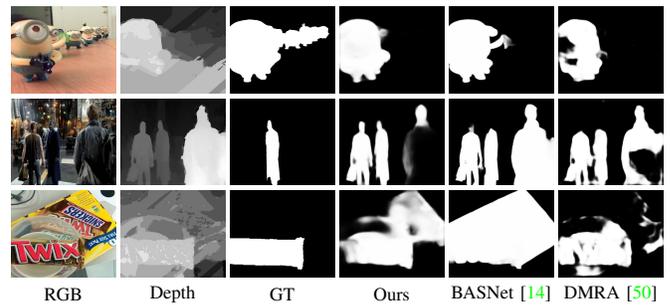

For future researchers to develop better algorithms, we provide some failure cases as shown in Fig. \ref{fig:failure}. It can be seen that it is difficult to locate salient objects perfectly in the following three aspects, whether it is based on the RGB SOD method (BASNet \cite{Qin_2019_CVPR}) or the RGB-D SOD method (Ours and DMRA \cite{Piao_2019_ICCV}): (1) Long-distance small and multiple salient objects. In the first image, when there are obvious differences in the size of salient targets in a scene, especially when the salient objects are multiple and small, and the depth map cannot provide the effective depth information of long-distance objects, it is difficult to completely detect all salient objects. (2) The conflict between the depth information and salient objects. In the second image, the highlighted object in the depth map (\ie, the close-range person on the right) is not the final salient object. For this case, it is difficult for the algorithm to suppress the interference of close-range objects, leading to the false alarms. (3) The complex and cluttered background regions. When confronting with the complex scenarios (\eg, the third image), especially when the depth map fails to provide accurate information, our algorithm is difficult to effectively extract the salient object from the complex background.

\section{Conclusion}
In this paper, we propose a novel framework \OURNET\ to achieve RGB-D SOD. Considering the contamination from the unreliable depth map, we model a saliency-orientated depth potentiality perception module to evaluate the potentiality of the depth map and weaken the contamination. To effectively aggregate the cross-modal complementarity, we propose a GMA module to highlight the saliency response and regulate the fusion rate of the cross-modal information. Finally, the multi-stage and multi-modality feature fusion are used to generate the discriminative RGB-D features and produce the saliency map.\ Experiments on eight RGB-D datasets demonstrate that the proposed network outperforms other $16$ state-of-the-art methods under different evaluation metrics.

\par
\ifCLASSOPTIONcaptionsoff
  \newpage
\fi
{
\bibliographystyle{IEEEtran}
\bibliography{egbib}

\begin{thebibliography}{10}
\providecommand{\url}[1]{#1}
\csname url@samestyle\endcsname
\providecommand{\newblock}{\relax}
\providecommand{\bibinfo}[2]{#2}
\providecommand{\BIBentrySTDinterwordspacing}{\spaceskip=0pt\relax}
\providecommand{\BIBentryALTinterwordstretchfactor}{4}
\providecommand{\BIBentryALTinterwordspacing}{\spaceskip=\fontdimen2\font plus
\BIBentryALTinterwordstretchfactor\fontdimen3\font minus
  \fontdimen4\font\relax}
\providecommand{\BIBforeignlanguage}[2]{{%
\expandafter\ifx\csname l@#1\endcsname\relax
\typeout{** WARNING: IEEEtran.bst: No hyphenation pattern has been}%
\typeout{** loaded for the language `#1'. Using the pattern for}%
\typeout{** the default language instead.}%
\else
\language=\csname l@#1\endcsname
\fi
#2}}
\providecommand{\BIBdecl}{\relax}
\BIBdecl

\bibitem{Survey}
W.~Wang, Q.~Lai, H.~Fu, J.~Shen, and H.~Ling, ``{Salient object detection in
  the deep learning era}: An in-depth survey,'' in \emph{arXiv preprint
  arXiv:1904.09146}, 2019.

\bibitem{REVIEW}
R.~Cong, J.~Lei, H.~Fu, M.-M. Cheng, W.~Lin, and Q.~Huang, ``Review of visual
  saliency detectioin with comprehensive information,'' \emph{IEEE Trans.
  Circuits Syst. Video Technol}, vol.~28, no.~10, pp. 2941--2959, 2019.

\bibitem{R1}
W.~Wang, J.~Shen, R.~Yang, and F.~Porikli, ``Saliency-aware video object
  segmentation,'' \emph{IEEE Trans. Pattern Anal. Mach. Intell.}, vol.~40,
  no.~1, pp. 20--33, 2018.

\bibitem{zhao2013unsupervised}
R.~Zhao, W.~Ouyang, and X.~Wang, ``Unsupervised salience learning for person
  re-identification,'' in \emph{CVPR}, 2013, pp. 3586--3593.

\bibitem{zhang2014saliency}
F.~Zhang, B.~Du, and L.~Zhang, ``Saliency-guided unsupervised feature learning
  for scene classification,'' \emph{IEEE Trans. Geosci. Remote Sens.}, vol.~53,
  no.~4, pp. 2175--2184, 2014.

\bibitem{R2}
W.~Wang, J.~Shen, Y.~Yu, and K.-L. Ma, ``Stereoscopic thumbnail creation via
  efficient stereo saliency detection,'' \emph{IEEE Trans. Vis. Comput. Graph},
  vol.~23, no.~8, pp. 2014--2027, 2017.

\bibitem{R3}
Q.~{Jiang}, F.~{Shao}, W.~{Lin}, K.~{Gu}, G.~{Jiang}, and H.~{Sun},
  ``Optimizing multistage discriminative dictionaries for blind image quality
  assessment,'' \emph{IEEE Trans. Multimedia}, vol.~20, no.~8, pp. 2035--2048,
  2018.

\bibitem{R4}
C.~Li, J.~Guo, R.~Cong, Y.~Pang, and B.~Wang, ``Underwater image enhancement by
  dehazing with minimum information loss and histogram distribution prior,''
  \emph{IEEE Trans. Image Process.}, vol.~25, no.~12, pp. 5664--5677, 2016.

\bibitem{krizhevsky2012imagenet}
A.~Krizhevsky, I.~Sutskever, and G.~E. Hinton, ``Imagenet classification with
  deep convolutional neural networks,'' in \emph{NeurIPS}, 2012, pp.
  1097--1105.

\bibitem{DSS}
Q.~Hou, M.-M. Cheng, X.~Hu, A.~Borji, Z.~Tu, and P.~H. Torr, ``Deeply
  supervised salient object detection with short connections,'' \emph{IEEE
  Trans. Pattern Anal. Mach. Intell.}, vol.~41, no.~4, pp. 815--828, 2019.

\bibitem{deng2018r3net}
Z.~Deng, X.~Hu, L.~Zhu, X.~Xu, J.~Qin, G.~Han, and P.-A. Heng,
  ``{R\textsuperscript{3}Net: Recurrent residual refinement network for
  saliency detection},'' in \emph{IJCAI}, 2018, pp. 684--690.

\bibitem{liu2018picanet}
N.~Liu, J.~Han, and M.-H. Yang, ``{PiCANet: Learning pixel-wise contextual
  attention for saliency detection},'' in \emph{CVPR}, 2018, pp. 3089--3098.

\bibitem{Liu2019PoolSal}
J.-J. Liu, Q.~Hou, M.-M. Cheng, J.~Feng, and J.~Jiang, ``A simple pooling-based
  design for real-time salient object detection,'' in \emph{CVPR}, 2019, pp.
  3917--3926.

\bibitem{Qin_2019_CVPR}
X.~Qin, Z.~Zhang, C.~Huang, C.~Gao, M.~Dehghan, and M.~Jagersand, ``{BASNet}:
  Boundary-aware salient object detection,'' in \emph{CVPR}, 2019, pp.
  7479--7489.

\bibitem{zhao2019EGNet}
J.-X. Zhao, J.-J. Liu, D.-P. Fan, Y.~Cao, J.~Yang, and M.-M. Cheng, ``{EGNet}:
  Edge guidance network for salient object detection,'' in \emph{ICCV}, 2019,
  pp. 8778--8787.

\bibitem{feng2019attentive}
M.~Feng, H.~Lu, and E.~Ding, ``Attentive feedback network for boundary-aware
  salient object detection,'' in \emph{CVPR}, 2019, pp. 1623--1632.

\bibitem{PFANet}
T.~Zhao and X.~Wu, ``Pyramid feature attention network for saliency
  detection,'' in \emph{CVPR}, 2019, pp. 3085--3094.

\bibitem{GCPANet}
Z.~Chen, Q.~Xu, R.~Cong, and Q.~Huang, ``Global context-aware progressive
  aggregation network for salient object detection,'' in \emph{AAAI}, 2020, pp.
  10\,599--10\,606.

\bibitem{Unsupervised}
J.~Zhang, T.~Zhang, Y.~Dai, M.~Harandi, and R.~Hartley, ``Deep unsupervised
  saliency detection: A multiple noisy labeling perspective,'' in \emph{CVPR},
  2018, pp. 9029--9038.

\bibitem{weaksupervised}
Y.~Zeng, Y.~Zhuge, H.~Lu, L.~Zhang, M.~Qian, and Y.~Yu, ``Multi-source weak
  supervision for saliency detection,'' in \emph{CVPR}, 2019, pp. 6074--6083.

\bibitem{wang2013saliency}
Q.~Wang, Y.~Yuan, P.~Yan, and X.~Li, ``Saliency detection by multiple-instance
  learning,'' \emph{IEEE Trans. Cybern.}, vol.~43, no.~2, pp. 660--672, 2013.

\bibitem{wang2018detecting}
Q.~Wang, M.~Chen, F.~Nie, and X.~Li, ``Detecting coherent groups in crowd
  scenes by multiview clustering,'' \emph{IEEE Trans. Pattern Anal. Mach.
  Intell.}, vol.~42, no.~1, pp. 46--58, 2018.

\bibitem{nips20}
Q.~Zhang, R.~Cong, J.~Hou, C.~Li, and Y.~Zhao, ``{CoADNet}: Collaborative
  aggregation-and-distribution networks for co-salient object detection,'' in
  \emph{Proc. NeurIPS}, 2020.

\bibitem{crmspl}
R.~Cong, J.~Lei, C.~Zhang, Q.~Huang, X.~Cao, and C.~Hou, ``Saliency detection
  for stereoscopic images based on depth confidence analysis and multiple cues
  fusion,'' \emph{IEEE Signal Process. Lett.}, vol.~23, no.~6, pp. 819--823,
  2016.

\bibitem{choi13iros_rgbdtracking}
C.~Choi and H.~I. Christensen, ``{RGB-D} object tracking: {A} particle filter
  approach on {GPU},'' in \emph{IROS}, 2013, pp. 1084--1091.

\bibitem{crmtip18}
R.~Cong, J.~Lei, H.~Fu, Q.~Huang, X.~Cao, and C.~Hou, ``Co-saliency detection
  for {RGBD} images based on multi-constraint feature matching and cross label
  propagation,'' \emph{IEEE Trans. Image Process.}, vol.~27, no.~2, pp.
  568--579, 2018.

\bibitem{crmtc19}
R.~Cong, J.~Lei, H.~Fu, W.~Lin, Q.~Huang, X.~Cao, and C.~Hou, ``An iterative
  co-saliency framework for {RGBD} images,'' \emph{IEEE Trans. Cybern.},
  vol.~49, no.~1, pp. 233--246, 2019.

\bibitem{ijcai20}
F.~Li, R.~Cong, H.~Bai, and Y.~He, ``{RGB-D} salient object detection with
  cross-modality modulation and selection,'' in \emph{Proc. IJCAI}, 2020, pp.
  534--543.

\bibitem{crmtmm19}
R.~Cong, J.~Lei, H.~Fu, Q.~Huang, X.~Cao, and N.~Ling, ``{HSCS}: Hierarchical
  sparsity based co-saliency detection for {RGBD} images,'' \emph{IEEE Trans.
  Multimedia}, vol.~21, no.~7, pp. 1660--1671, 2019.

\bibitem{crmtc20}
R.~Cong, J.~Lei, H.~Fu, J.~Hou, Q.~Huang, and S.~Kwong, ``Going from {RGB} to
  {RGBD} saliency: A depth-guided transformation model,'' \emph{IEEE Trans.
  Cybern.}, vol.~50, no.~8, pp. 3627--3639, 2020.

\bibitem{lcy2020tc}
C.~Li, R.~Cong, S.~Kwong, J.~Hou, H.~Fu, G.~Zhu, D.~Zhang, and Q.~Huang,
  ``{ASIF-Net}: Attention steered interweave fusion network for {RGB-D} salient
  object detection,'' \emph{IEEE Trans. Cybern.}, vol.~PP, no.~99, pp. 1--13,
  2020.

\bibitem{eccv20}
C.~Li, R.~Cong, Y.~Piao, Q.~Xu, and C.~C. Loy, ``{RGB-D} salient object
  detection with cross-modality modulation and selection,'' in \emph{Proc.
  ECCV}, 2020, pp. 1--17.

\bibitem{zhao2019contrast}
J.-X. Zhao, Y.~Cao, D.-P. Fan, M.-M. Cheng, X.-Y. Li, and L.~Zhang, ``Contrast
  prior and fluid pyramid integration for {RGBD} salient object detection,'' in
  \emph{CVPR}, 2019, pp. 3927--3936.

\bibitem{otsu1979threshold}
N.~Otsu, ``A threshold selection method from gray-level histograms,''
  \emph{IEEE Trans. Syst. Man Cybern.}, vol.~9, no.~1, pp. 62--66, 1979.

\bibitem{gupta2014learning}
S.~Gupta, R.~Girshick, P.~Arbel{\'a}ez, and J.~Malik, ``{Learning rich features
  from RGB-D images for object detection and segmentation},'' in \emph{ECCV},
  2014, pp. 345--360.

\bibitem{krahenbuhl2011efficient}
P.~Kr{\"a}henb{\"u}hl and V.~Koltun, ``Efficient inference in fully connected
  crfs with gaussian edge potentials,'' in \emph{NeurIPS}, 2011, pp. 109--117.

\bibitem{DSR}
X.~Li, H.~Lu, L.~Zhang, X.~Ruan, and M.-H. Yang, ``Saliency detection via dense
  and sparse reconstruction,'' in \emph{ICCV}, 2013, pp. 2976--2983.

\bibitem{RBD}
W.~Zhu, S.~Liang, Y.~Wei, and J.~Sun, ``Saliency optimization from robust
  background detection,'' in \emph{CVPR}, 2014, pp. 2814--2821.

\bibitem{DCLC}
L.~Zhou, Z.~Yang, Q.~Yuan, Z.~Zhou, and D.~Hu, ``Salient region detection via
  integrating diffusion-based compactness and local contrast,'' \emph{IEEE
  Trans. Image Process.}, vol.~24, no.~11, pp. 3308--3320, 2015.

\bibitem{SMD}
H.~Peng, B.~Li, H.~Ling, W.~Hua, W.~Xiong, and S.~Maybank, ``Salient object
  detection via structured matrix decomposition,'' \emph{IEEE Trans. Pattern
  Anal. Mach. Intell.}, vol.~39, no.~4, pp. 818--832, 2017.

\bibitem{peng2014rgbd}
H.~Peng, B.~Li, W.~Xiong, W.~Hu, and R.~Ji, ``{RGBD salient object detection: A
  benchmark and algorithms},'' in \emph{ECCV}, 2014, pp. 92--109.

\bibitem{song2017depth}
H.~Song, Z.~Liu, H.~Du, G.~Sun, O.~Le~Meur, and T.~Ren, ``Depth-aware salient
  object detection and segmentation via multiscale discriminative saliency
  fusion and bootstrap learning,'' \emph{IEEE Trans. Image Process.}, vol.~26,
  no.~9, pp. 4204--4216, 2017.

\bibitem{feng2016local}
D.~Feng, N.~Barnes, S.~You, and C.~McCarthy, ``{Local background enclosure for
  RGB-D salient object detection},'' in \emph{CVPR}, 2016, pp. 2343--2350.

\bibitem{ju2014depth}
R.~Ju, L.~Ge, W.~Geng, T.~Ren, and G.~Wu, ``Depth saliency based on anisotropic
  center-surround difference,'' in \emph{ICIP}, 2014, pp. 1115--1119.

\bibitem{fan2014salient}
X.~Fan, Z.~Liu, and G.~Sun, ``Salient region detection for stereoscopic
  images,'' in \emph{DSP}, 2014, pp. 454--458.

\bibitem{cheng2014depth}
Y.~Cheng, H.~Fu, X.~Wei, J.~Xiao, and X.~Cao, ``Depth enhanced saliency
  detection method,'' in \emph{ICIMCS}, 2014, p.~23.

\bibitem{qu2017rgbd}
L.~Qu, S.~He, J.~Zhang, J.~Tian, Y.~Tang, and Q.~Yang, ``{RGBD salient object
  detection via deep fusion},'' \emph{IEEE Trans. Image Process.}, vol.~26,
  no.~5, pp. 2274--2285, 2017.

\bibitem{zhu2019pdnet}
C.~Zhu, X.~Cai, K.~Huang, T.~H. Li, and G.~Li, ``{PDNet}: Prior-model guided
  depth-enhanced network for salient object detection,'' in \emph{ICME}, 2019,
  pp. 199--204.

\bibitem{chen2018progressively}
H.~Chen and Y.~Li, ``{Progressively complementarity-aware fusion network for
  RGB-D salient object detection},'' in \emph{CVPR}, 2018, pp. 3051--3060.

\bibitem{Piao_2019_ICCV}
Y.~{Piao}, W.~{Ji}, J.~{Li}, M.~{Zhang}, and H.~{Lu}, ``Depth-induced
  multi-scale recurrent attention network for saliency detection,'' in
  \emph{ICCV}, 2019, pp. 7253--7262.

\bibitem{fan2019D3Net}
D.-P. Fan, Z.~Lin, Z.~Zhang, M.~Zhu, and M.-M. Cheng, ``{Rethinking RGB-D}
  salient object detection: Models, data sets, and large-scale benchmarks,''
  \emph{IEEE Trans. Neural Netw. Learn. Syst.}, 2020.

\bibitem{han2017cnns}
J.~Han, H.~Chen, N.~Liu, C.~Yan, and X.~Li, ``{CNNs-based RGB-D saliency
  detection via cross-view transfer and multiview fusion},'' \emph{IEEE Trans.
  Cybern.}, vol.~48, no.~11, pp. 3171--3183, 2017.

\bibitem{achanta2009frequency}
R.~Achanta, S.~Hemami, F.~Estrada, and S.~S{\"u}sstrunk, ``Frequency-tuned
  salient region detection,'' in \emph{CVPR}, 2009, pp. 1597--1604.

\bibitem{vaswani2017attention}
A.~Vaswani, N.~Shazeer, N.~Parmar, J.~Uszkoreit, L.~Jones, A.~N. Gomez,
  {\L}.~Kaiser, and I.~Polosukhin, ``Attention is all you need,'' in
  \emph{NeurIPS}, 2017, pp. 5998--6008.

\bibitem{wang2018non}
X.~Wang, R.~Girshick, A.~Gupta, and K.~He, ``Non-local neural networks,'' in
  \emph{CVPR}, 2018, pp. 7794--7803.

\bibitem{girshick2015fast}
R.~Girshick, ``{Fast R-CNN},'' in \emph{ICCV}, 2015, pp. 1440--1448.

\bibitem{niu2012leveraging}
Y.~Niu, Y.~Geng, X.~Li, and F.~Liu, ``Leveraging stereopsis for saliency
  analysis,'' in \emph{CVPR}, 2012, pp. 454--461.

\bibitem{li2014saliency}
N.~Li, J.~Ye, Y.~Ji, H.~Ling, and J.~Yu, ``Saliency detection on light field,''
  in \emph{CVPR}, 2014, pp. 2806--2813.

\bibitem{zhu2017three}
C.~Zhu and G.~Li, ``A three-pathway psychobiological framework of salient
  object detection using stereoscopic technology,'' in \emph{ICCV}, 2017, pp.
  3008--3014.

\bibitem{fan2017structure}
D.-P. Fan, M.-M. Cheng, Y.~Liu, T.~Li, and A.~Borji, ``{Structure-measure: A
  new way to evaluate foreground maps},'' in \emph{ICCV}, 2017, pp. 4548--4557.

\bibitem{he2016deep}
K.~He, X.~Zhang, S.~Ren, and J.~Sun, ``Deep residual learning for image
  recognition,'' in \emph{CVPR}, 2016, pp. 770--778.

\bibitem{wang2019adaptive}
N.~Wang and X.~Gong, ``Adaptive fusion for {RGB-D} salient object detection,''
  \emph{IEEE Access}, vol.~7, pp. 55\,277--55\,284, 2019.

\bibitem{chen2019three}
H.~Chen and Y.~Li, ``{Three-stream attention-aware network for RGB-D salient
  object detection},'' \emph{IEEE Trans. Image Process.}, vol.~28, no.~6, pp.
  2825--2835, 2019.

\bibitem{chen2019multi}
H.~Chen, Y.~Li, and D.~Su, ``{Multi-modal fusion network with multi-scale
  multi-path and cross-modal interactions for RGB-D salient object
  detection},'' \emph{Pattern Recognition}, vol.~86, pp. 376--385, 2019.

\bibitem{shigematsu2017learning}
R.~Shigematsu, D.~Feng, S.~You, and N.~Barnes, ``Learning {RGB-D} salient
  object detection using background enclosure, depth contrast, and top-down
  features,'' in \emph{ICCV}, 2017, pp. 2749--2757.

\end{thebibliography}
}

\end{document}